\documentclass[conference]{IEEEtran}
\IEEEoverridecommandlockouts
\usepackage{cite}
\usepackage{amsmath,amssymb,amsfonts}
\usepackage{amsthm}
\usepackage{algorithmic}
\usepackage{setspace}
\usepackage{algorithm}
\usepackage{graphicx}
\usepackage{enumerate}
\usepackage{subcaption}
\usepackage{multirow}
\usepackage{booktabs}
\usepackage{textcomp}
\usepackage{xcolor}
\def\BibTeX{{\rm B\kern-.05em{\sc i\kern-.025em b}\kern-.08em
    T\kern-.1667em\lower.7ex\hbox{E}\kern-.125emX}}
\begin{document}
\newcommand{\jong}[1]{\textcolor{blue}{#1}}
\newcommand{\hoon}[1]{\textcolor{blue}{#1}}

\newcommand{\algname}{HypHGT}

\theoremstyle{definition}
\newtheorem{definition}{Definition}

\title{Hyperbolic Heterogeneous Graph Transformer}

% \author{Jongmin Park\textsuperscript{1}, Seunghoon Han\textsuperscript{1}, Won-Yong Shin\textsuperscript{2},  Sungsu Lim\textsuperscript{1}\textsuperscript{*}}

\iffalse
\author{Jongmin Park$^{*}$\thanks{\IEEEauthorrefmark{1}Equally contributed.}, Seunghoon Han$^{*}$, Hyewon Lee, Won-Yong Shin,~\IEEEmembership{Senior Member,~IEEE}, Sungsu Lim$^{\dagger}$\thanks{\IEEEauthorrefmark{2}Corresponding author.},~\IEEEmembership{Member,~IEEE}}

\author{Jongmin Park$^{*}$\thanks{\IEEEauthorrefmark{1}Equally contributed.}, Seunghoon Han$^{*}$, Hyewon Lee, Won-Yong Shin, Sungsu Lim$^{\dagger}$\thanks{\IEEEauthorrefmark{2}Corresponding author.}}
\fi

\author{
\IEEEauthorblockN {
   Jongmin Park$^{1, *}$\thanks{\IEEEauthorrefmark{1}Equally contributed.},       %%% instead of \IEEEauthorrefmark{1},
    Seunghoon Han$^{1, *}$,
    Hyewon Lee$^{1},$%%% instead of \IEEEauthorrefmark{1}
    Won-Yong Shin$^{2}$,
    Sungsu Lim$^{1, \dagger}$\thanks{\IEEEauthorrefmark{2}Corresponding author.}
}

\IEEEauthorblockA {
  $^{1}$Department Computer Science and Engineering, Chungnam National University, Daejeon, South Korea \\
  %\{pa5398, tmdgns129\}@g.cnu.ac.kr, sungsu@cnu.ac.kr
}
\IEEEauthorblockA {
    $^{2}$School of Mathematics and Computing (Computational Science and Engineering), Yonsei University, Seoul, South Korea \\
}

\IEEEauthorblockA {
    \{pa5398, tmdgns129, noweyh927\}@g.cnu.ac.kr, wy.shin@yonsei.ac.kr, sungsu@cnu.ac.kr
}
}

\maketitle

\begin{abstract}
In heterogeneous graphs, we can observe complex structures such as tree-like or hierarchical structures. Recently, the hyperbolic space has been widely adopted in many studies to effectively learn these complex structures. Although these methods have demonstrated the advantages of the hyperbolic space in learning heterogeneous graphs, most existing methods still have several challenges. They rely heavily on tangent-space operations, which often lead to mapping distortions during frequent transitions. Moreover, their message-passing architectures mainly focus on local neighborhood information, making it difficult to capture global hierarchical structures and long-range dependencies between different types of nodes. To address these limitations, we propose Hyperbolic Heterogeneous Graph Transformer (\algname{}), which effectively and efficiently learns heterogeneous graph representations entirely within the hyperbolic space. Unlike previous message-passing based hyperbolic heterogeneous GNNs, \algname{} naturally captures both local and global dependencies through transformer-based architecture. Furthermore, the proposed relation-specific hyperbolic attention mechanism in \algname{}, which operates with linear time complexity, enables efficient computation while preserving the heterogeneous information across different relation types. This design allows \algname{} to effectively capture the complex structural properties and semantic information inherent in heterogeneous graphs. We conduct comprehensive experiments to evaluate the effectiveness and efficiency of HypHGT, and the results demonstrate that it consistently outperforms state-of-the-art methods in node classification task, with significantly reduced training time and memory usage.
%In heterogeneous graphs, we can observe complex structures such as tree-like or hierarchical structures. Recently, the hyperbolic space has been widely adopted in many studies to effectively learn these complex structures. Despite the effectiveness of the hyperbolic space in modeling such structures, previous works still have some limitations: they rely heavily on the tangent space to perform hyperbolic operations that causes mapping distortions, incur substantial computational overhead due to complex hyperbolic operations, and require dataset-specific metapath sampling that depend on prior domain knowledge. Moreover, their focus on local aggregation makes it difficult to capture the global hierarchical structures of heterogeneous graphs.

%To overcome these limitations, we proposed a novel Hyperbolic Heterogeneous Graph Transformer (\algname{}), which effectively and efficiently learns heterogeneous graph representations entirely within the hyperbolic space. Specifically, for computational efficiency, \algname{} performs linear attention entirely within the hyperbolic space corresponding to each relation type. Moreover, to effectively capture the structural properties specific to each relation type, we use a distinct hyperbolic space for each of them. We conduct comprehensive experiments to evaluate the effectiveness and efficiency of \algname{}. The results show that \algname{} outperforms state-of-the-art methods in node classification tasks, while requiring less computational time and memory consumption.

\end{abstract}

\section{Introduction}
\IEEEPARstart{H}{eterogeneous graphs}, which consist of multiple types of nodes and links, effectively represent various real-world scenarios such as academic networks~\cite{zhang2019heterogeneous, ji2021heterogeneous, liang2022meta}, social networks~\cite{dong2012link,qiao2020heterogeneous,salamat2021heterographrec, 10004751}, and molecular structures~\cite{shui2020heterogeneous,ji2023metapath, jiang2023pharmacophoric}. In heterogeneous graphs, we can observe various complex structures such as tree-like or hierarchical structures. However, representing such complex structures in the Euclidean space is challenging due to its limited ability to capture the hierarchical and power-law characteristics inherent in heterogeneous graphs. Therefore, it is necessary to use an embedding space that can naturally capture such structural patterns beyond the limitations of the Euclidean space~\cite{pei2020curvature, balazevic2019multi, nickel2017poincare}.
% For example, as illustrated in Figure~\ref{figure:toy_example}, in a movie network with hierarchical structures, distortion can occur when two nodes (Actor 1 and Actor 2) that are far apart in the geodesic distance are represented in close proximity within the Euclidean space. Since the Euclidean space expands polynomially, it struggles to preserve exponentially growing hierarchical structures. In contrast, the hyperbolic space expands exponentially from a north pole, allowing it to naturally capture such hierarchical and tree-like structures.

The hyperbolic space, which has a constant negative curvature, expands exponentially from a north pole. This property enables it to represent hierarchical and complex structures more effectively than Euclidean space~\cite{wang2019hyperbolic, pan2021hyperbolic, wang2021knowledge, 10361607, 9999499}.
Due to this characteristic of the hyperbolic space, many recent studies~\cite{liu2019hyperbolic, chami2019hyperbolic, zhang2021hyperbolic, wang2019hyperbolic, li2023multi, park2024hyperbolic, park2024multi} have used the hyperbolic space as an embedding space to effectively capture such complex structures. SHAN~\cite{li2023multi} proposed hyperbolic heterogeneous Graph Neural Network (GNN) that captures complex structures within sampled simpicial complexes from heterogeneous graphs and uses graph attention mechanisms in the hyperbolic space to learn multi-order relations. HHGAT~\cite{park2024hyperbolic} proposed hyperbolic heterogeneous graph attention networks to learn semantic information and complex structures by leveraging metapaths, which are defined as the ordered sequence of node and link types that represent semantic structures within heterogeneous graphs. MSGAT~\cite{park2024multi} further extended HHGAT by using multiple hyperbolic spaces to effectively learn diverse power-law structures corresponding to different metapaths.

Although existing models have achieved remarkable performance in heterogeneous graph representation learning within the hyperbolic space, they still suffer from several challenges: 

\begin{itemize}
    \item \textbf{Challenge 1} (\textbf{Over-frequent mapping}). Previous hyperbolic heterogeneous GNNs heavily rely on the tangent space to perform hyperbolic graph convolution operations, which often leads to mapping distortions during exponential and logarithmic mappings between the hyperbolic and tangent spaces. 
    \item \textbf{Challenge 2} (\textbf{Prior-knowledge requirement}). They rely on predefined structures sampled from heterogeneous graphs, such as metapath instances, which carefully defined based on domain-specific knowledge for each dataset, thereby restricting generalization and adapt to unseen relational patterns.
    \item \textbf{Challenge 3} (\textbf{High computational cost}). Due to the use of complex hyperbolic operations and structure-based sampling processes, these models often require substantial computational costs in terms of both time and memory during training.
    \item \textbf{Challenge 4} (\textbf{Absence of global information}). Since these models~\cite{park2024hyperbolic, park2024multi, li2023multi} only aggregate information from local neighbors, they struggle to capture the global hierarchical structures inherent in heterogeneous graphs.
\end{itemize}

\begin{figure}[t!]
\centering    
\includegraphics[width=\columnwidth]{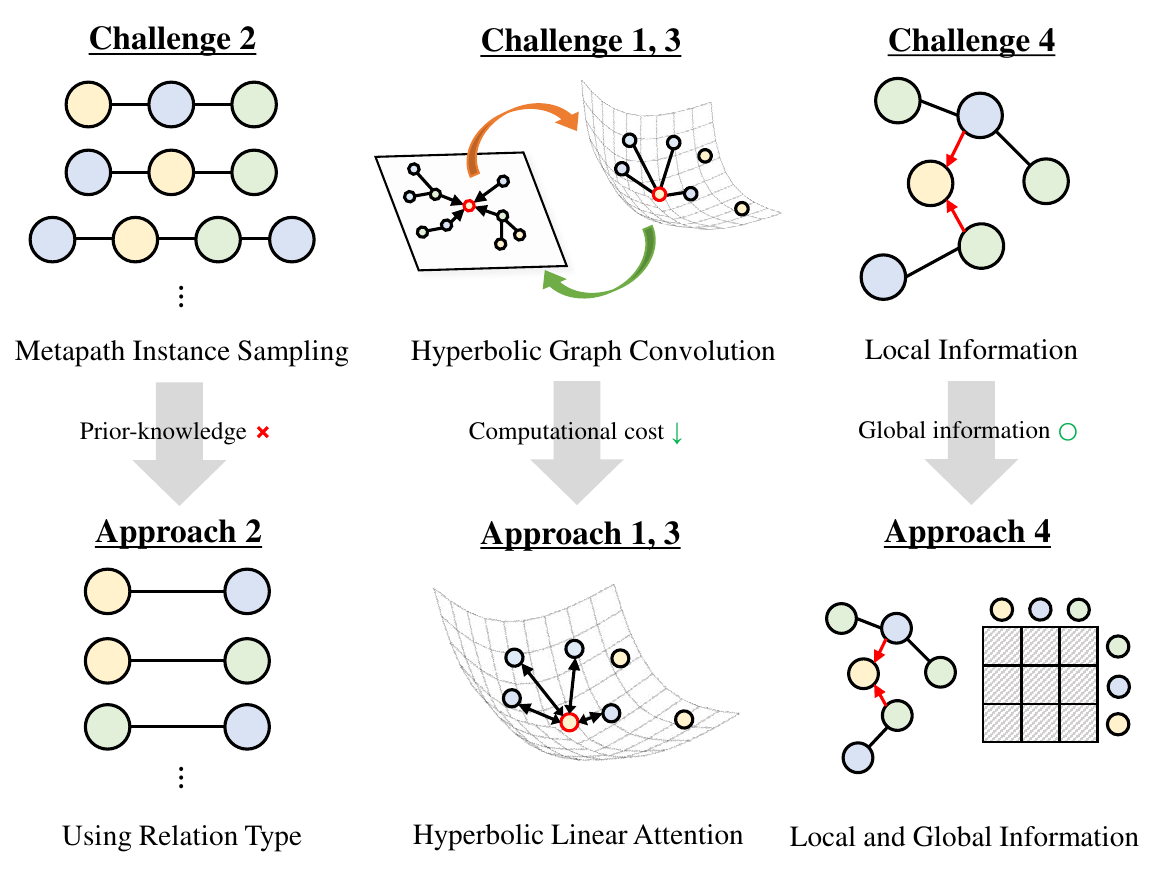}
    \caption{Challenges (top) and corresponding approaches (bottom) in previous hyperbolic heterogeneous GNNs addressed by the proposed \algname{}.}
    \label{figure:toy_example}
\end{figure}

Figure~\ref{figure:toy_example} illustrates the main challenges in previous hyperbolic heterogeneous GNNs and the corresponding approaches proposed in this work.
Challenges 1 and 3 arise from frequent tangent–hyperbolic mappings and the high computational costs of hyperbolic graph convolution, Challenge 2 stems from dependence on predefined metapaths, and Challenge 4 reflects the inability to capture global information due to local neighbor information aggregation.
\algname{} addresses these challenges by performing efficient operations entirely within the hyperbolic space through linear self-attention (Approach 1 and 3), replacing predefined metapaths with relation-type modeling (Approach 2), and capturing both local and global dependencies via transformer-based architecture (Approach 4).
The details of these approaches are described as follows.

%To overcome these limitations, we propose a novel Hyperbolic Heterogeneous Graph Transformer (\algname{}), which is designed to comprehensively address the above challenges as follows:

\textbf{Approach 1}. \algname{} learns heterogeneous graph representations entirely within the hyperbolic space, avoiding unnecessary tangent-hyperbolic mappings except for the initialization of node features and the final projection required for downstream tasks. By performing all operations directly in the hyperbolic space, \algname{} effectively reduces mapping distortions during the exponential and logarithmic mappings between the hyperbolic and tangent spaces.

\textbf{Approach 2}. We propose a relation-aware hyperbolic attention mechanism that learn distinct relation-specific structures within relation-specific hyperbolic spaces. This enables \algname{} to effectively capture semantic heterogeneity across different link types and to learn diverse complex structures without relying on predefined structures such as metapath instances. 
% relation 기반의 sampling은 prio-knowledge가 필요 없다는 내용도 들어가면 좋을듯

\textbf{Approach 3}. To address the high computational cost caused by softmax-based self-attention mechanisms in transformer architectures, we use a linear attention mechanism defined directly in the hyperbolic space. It can reduce both computational time and memory consumption.

\textbf{Approach 4}. 
% To effectively capture both local and global structural information, 
\algname{} integrates hyperbolic transformer layers that learn global information with heterogeneous GNN layers that aggregate local neighbor information. This architecture enables \algname{} to learn comprehensive representations that capture both local relationships and global hierarchical structures inherent in heterogeneous graphs.

Our study is motivated by the empirical observation that message-passing-based hyperbolic heterogeneous GNNs degrade substantially on sparse heterogeneous graphs, due to limited local neighborhood information.

\begin{figure}[t!]
\captionsetup[subfigure]{justification=centering}
\centering
    \begin{minipage}[b]{0.49\columnwidth}
        \centering
        \includegraphics[width=\linewidth]{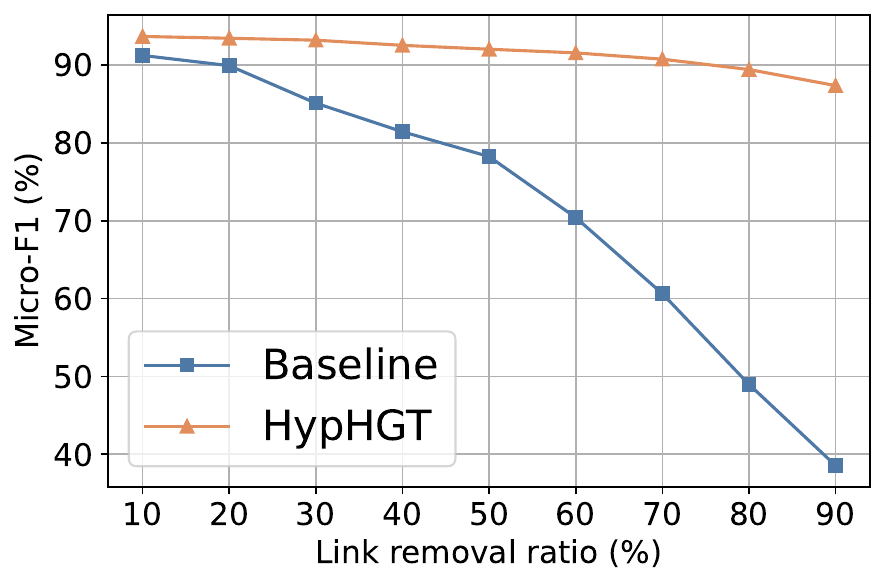}
        \subcaption{}
    \end{minipage}
    \begin{minipage}[b]{0.49\linewidth}
        \centering
        \includegraphics[width=\linewidth]{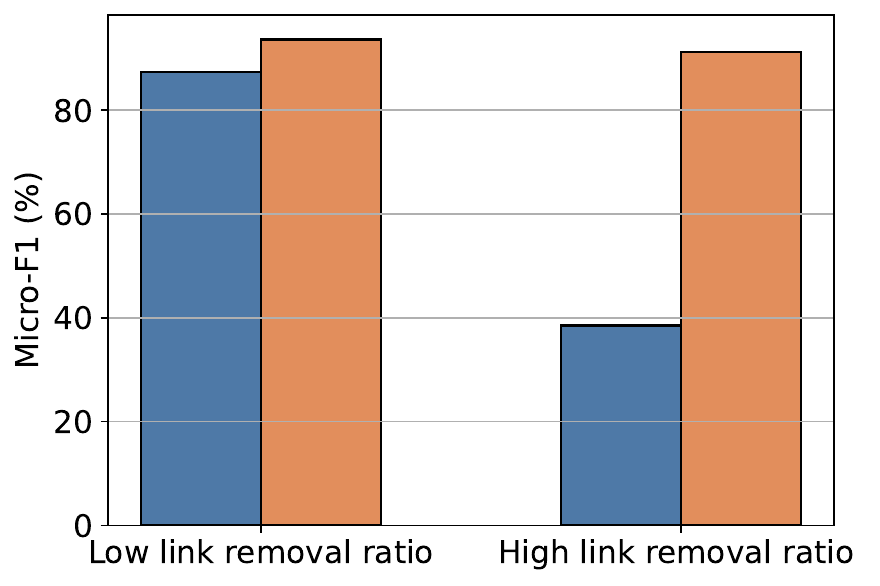}
        \subcaption{}
    \end{minipage}
\caption{Comparison between a message-passing-based hyperbolic heterogeneous GNN and HypHGT under increasing graph sparsity. (a) Node classification performance on the ACM dataset under increasing link removal ratios. (b) Node classification performance comparison under high- and low-sparsity levels.}
\label{figure:comparison}
\end{figure}

Figure~\ref{figure:comparison} illustrates a comparison between message-passing-based hyperbolic heterogeneous GNN with \algname{} under increasing graph sparsity. Although baseline model uses the hyperbolic space, their reliance on local neighborhood aggregation limits their ability to effectively capture global hierarchical structures. As graph sparsity increases, the neighborhood information becomes limited, leading to rapid performance degradation. In contrast, \algname{} leverages a transformer-based architecture to directly capture long-range dependencies, enabling more robust learning global hierarchical structures even when local connectivity is severely reduced.

In summary, the transformer-based architecture allows \algname{} to overcome the limitations of message-passing-based hyperbolic heterogeneous GNNs by capturing long-range dependencies and modeling global hierarchical structures beyond local neighborhoods.
Moreover, due to the exponential growth of the hyperbolic space with a negative curvature, nodes at different hierarchical levels can be embedded with geometrically consistent separations, enabling the transformer-based architecture to better capture and preserve their hierarchical structural properties.
Furthermore, since \algname{} uses a linear self-attention mechanism, it achieves efficient heterogeneous graph learning while preserving the expressive power required to represent complex structures.

The main contributions of our work can be summarized as follows:
\begin{itemize}
    \item We propose a novel model for learning heterogeneous graph representations, \algname{}. To the best of our knowledge, \algname{} is the first transformer-based embedding model for heterogeneous graphs in the hyperbolic space.
    \item We conduct comprehensive experiments to evaluate the effectiveness and efficiency of \algname{}. The experimental results demonstrate that \algname{} outperforms state-of-the-art methods in various downstream tasks with heterogeneous graphs.
    \item We analyze the impact of hyperbolic spaces on the \algname{} performance, and observe that the learnable curvatures in relation-specific hyperbolic spaces are adaptively optimized according to the structural distributions of relation types.
\end{itemize}
The remainder of this paper is organized as follows. In Section~\ref{sec:related work}, we provide an overview of relevant works. In Section~\ref{sec:Preliminaries}, we introduce the preliminaries and theoretical background. In Section~\ref{sec:Methodology}, we present the proposed \algname{} in detail. In Section~\ref{sec:Experiments}, we provide experimental results and in-depth analyses. Finally, in Section~\ref{sec:Conclusion}, we conclude the paper with a summary of key findings. Table~\ref{tab:notation} summarizes notations used throughout the paper.

\section{Related work}
\algname{} is related to three broader areas of research, namely Euclidean heterogeneous GNNs, hyperbolic heterogeneous GNNs, and hyperbolic transformer.
\label{sec:related work}
\subsection{Euclidean Heterogeneous GNNs}
GNNs~\cite{wu2020comprehensive, zhou2020graph, khemani2024review, 10818675} have demonstrated effectiveness for graph representation learning, and recent research has focused on extending them to heterogeneous graphs to learn heterogeneous information within such graphs. Most recent studies leverage metapaths to capture semantic information inherent in heterogeneous graphs.

HAN~\cite{wang2019heterogeneous} introduced a hierarchical graph attention network that aggregates information at two different levels: node-level and semantic-level attention. At the node level, HAN aggregates information from metapath-based neighbors, while at the semantic level, it aggregates the semantic information learned from different metapaths. However, during node-level attention, the intermediate nodes within metapath instances are not considered.
To address this limitation, MAGNN~\cite{fu2020magnn} extended HAN by considering intermediate nodes into the learning process, thereby enhancing the ability to capture richer semantic information within metapaths.
Nevertheless, both HAN and MAGNN require predefined metapaths to learn semantic information inherent in heterogeneous graphs, and defining appropriate metapaths without sufficient domain knowledge remains a challenging task across different datasets.

To address this challenge, recent studies proposed automatic metapath selection. GTN~\cite{yun2019graph} introduced graph transformer layers that learn a soft selection of link types and composite relations to automatically generate informative metapaths. GraphMSE~\cite{li2021graphmse} further proposed an automatic metapath selection framework based on semantic feature space alignment.

In contrast, several heterogeneous GNNs do not rely on predefined metapath construction. Instead, they employ link-type dependent graph convolution operations to implicitly capture semantic information within heterogeneous graphs. HGT~\cite{hu2020heterogeneous} introduced node- and link-type dependent attention mechanisms to model diverse relation types in heterogeneous graphs, while Simple-HGN~\cite{lv2021we} extended graph attention networks by incorporating link-type information into the calculation of attention scores.

Although aforementioned Euclidean heterogeneous GNNs have achieved remarkable progress in heterogeneous graph representation learning, they still struggle to effectively capture the hierarchical structures commonly observed in such graphs.

\subsection{Hyperbolic Heterogeneous GNNs}
Since Euclidean space inherently has limitations in effectively representing the complex structures within heterogeneous graphs, recent studies have proposed hyperbolic heterogeneous GNNs that use the hyperbolic space as an embedding space. SHAN~\cite{li2023multi} proposed a simplicial hyperbolic attention network to capture complex structural patterns within sampled simplicial complexes, which represent multi-order relations inherent in heterogeneous graphs. HHGAT~\cite{park2024hyperbolic} used the hyperbolic space to capture power-law structures based on metapaths, enabling effective learning of semantic information within heterogeneous graphs. Although HHGAT demonstrated remarkable performance in heterogeneous graph representation learning, relying solely on a single hyperbolic space presents limitations when learning the diverse power-law structures within heterogeneous graphs.
To overcome this limitation, MSGAT~\cite{park2024multi} extended HHGAT by using multiple hyperbolic spaces to effectively capture diverse power-law structures corresponding to different metapaths.

Previous hyperbolic heterogeneous GNNs have demonstrated the effectiveness of the hyperbolic space in learning complex structures within heterogeneous graphs. However, they suffer from mapping distortions caused by frequent transitions between the hyperbolic and tangent spaces. Their complex hyperbolic operations lead to increased computational costs, such as computatinal time and memory consumption.

\subsection{Hyperbolic Transformer}
Transformers~\cite{vaswani2017attention, lin2022survey, 9609539} have recently been extended to the hyperbolic space, to enhance their ability to capture complex structures that are difficult to capture in Euclidean space. The hyperbolic space, with its constant negative curvature and exponential expansion can provide a natural representation for complex structures. Due to these properties, various hyperbolic Transformer models~\cite{11003808, tang2025hyperrole, ermolov2022hyperbolic} have been proposed. Hyperbolic Transformer~\cite{gulcehre2018hyperbolic} proposed a hyperbolic attention network that uses hyperbolic distance to compute attention scores and performs hyperbolic aggregation through the Einstein midpoint to obtain representations in the hyperbolic space. Similarly, HYBONET~\cite{chen2022fully} computes attention scores based on the Lorentzian distance and performs hyperbolic aggregation using the Lorentz centroid. Since the self-attention mechanism in these models has quadratic time complexity, their scalability is limited. To address this limitation, Hypformer~\cite{yang2024hypformer} proposed a hyperbolic transformer architecture with linear-time complexity to ensure scalability. Moreover, all computations and representation learning in Hypformer are performed entirely within the fully hyperbolic space, reducing distortions caused by frequent mappings between the hyperbolic and tangent spaces. 

In this work, we extend the fully hyperbolic Transformer architecture with linear-time complexity to heterogeneous graphs, enabling effective and efficient heterogeneous graph representation learning entirely within the hyperbolic space.

\begin{table}[t!]
\centering
\caption{Notations.}
\label{tab:notation}
\resizebox{\columnwidth}{!}{%
\begin{tabular}{cl}
\hline
\multicolumn{1}{l}{Notation} & Explanation \\ \hline
$\mathcal{G}$ & A Heterogeneous graph \\
$\epsilon$ & A relation type ($\forall\epsilon \in{T}_\mathcal{E}$) \\
${T}_\mathcal{E}$ & The set of relation types \\
$\mathbb{L}^{n,c}$ & $n$-dimensional Lorentz model with curvature $c\;(c<0)$ \\
$\mathbb{R}^n$ & $n$-dimensional Euclidean space \\
$\mathcal{T}_\mathbf{x}\textbf{}\mathbb{L}^{n,c}$ & Tangent space at point $\mathbf{x}$ \\
$\text{exp}^c_\mathbf{x}(\cdot)$ & Exponential map at point $\mathbf{x}$, $\text{exp}^c_\mathbf{x}:\mathcal{T}_\mathbf{x}\mathbb{L}^{n,c}\rightarrow\mathbb{L}^{n,c}$ \\
$\text{log}^c_\mathbf{x}(\cdot)$ & Logarithmic map at point $\mathbf{x}$, $\text{log}^c_\mathbf{x}:\mathbb{L}^{n,c}\rightarrow \mathcal{T}_\mathbf{x}\mathbb{L}^{n,c}$ \\
$\text{HT($\cdot$)}$ & Hyperbolic transformation \\
HR($\cdot$) & Hyperbolic readjustment and refinement \\
$\phi(\cdot)$ & Kernel-function for linear-attention\\
\hline
\end{tabular}%
}
\end{table}

\section{Preliminaries}
In this section, we introduce key concepts related to heterogeneous graphs and the Lorentz model adopted in this work.
\label{sec:Preliminaries}
\subsection{Heterogeneous Graph}
\begin{definition}[\bf Heterogeneous graph]
A heterogeneous graph~\cite{wang2019heterogeneous} is defined as a graph $\mathcal{G} = (\mathcal{V}, \mathcal{E}, f_v(\cdot), f_e(\cdot))$. $\mathcal{V}$ is a set of nodes, $\mathcal{E}$ is a set of links, $f_v(\cdot) :\;\mathcal{V} \rightarrow T_\mathcal{V}$ is a node type mapping function, and $f_e(\cdot) :\;\mathcal{E} \rightarrow T_\mathcal{E}$ is a link type mapping function, where $T_\mathcal{V}$ and $\mathcal{T}_\mathcal{E}$ are sets of node types and link types, respectively, with $|T_\mathcal{V}| + |T_\mathcal{E}| > 2$.
\end{definition}

\subsection{Lorentz model}

In hyperbolic geometry, the Lorentz model~\cite{chen2022fully, yang2024hypformer} is widely adopted for its numerical stability and computational efficiency in performing geometric operations. The mathematical formulations underlying the Lorentz model are briefly described in Definitions 2-6.

\begin{definition}[\bf Lorentz Model] An $n$-dimensional Lorentz model with negative curvature $c\;(c<0)$ is defined by Riemannian manifold $\mathbb{L}^{n,c} = (\mathcal{L}^n, \mathbf{g}^c)$. $\mathcal{L}^n$ is the upper sheet of hyperboloid (hyper-surface) $\mathbf{g}^c=\text{diag}(1/c, 1, \cdots, 1)$ is the Riemmanian metric tensor. Each point in $\mathbb{L}^{n,c}$ can be represented as $\mathbf{x}=\begin{bmatrix} x_t\\ \mathbf{x}_s\\ \end{bmatrix}$, where $\mathbf{x}\in\mathbb{R}^{n+1}$, $x_t\in\mathbb{R}$, and $\mathbf{x}_s\in\mathbb{R}^{n}$. The set of points $\mathbb{L}^{n,c}$ is defined as follow:
\begin{equation}
    \mathbb{L}^{n,c} := \{\mathbf{x}\in\mathbb{R}^{n+1}\;|\;\langle\mathbf{x},\mathbf{x}\rangle_\mathcal{L}=1/c, \;x_t>0\}.
\end{equation}
Here, $\langle\mathbf{x},\;\mathbf{y}\rangle_{\mathcal{L}} = -x_t y_t\;+\mathbf{x}_s^{T}\mathbf{y}_s=\mathbf{x}^{T}\mathbf{g}^c\mathbf{y}$ is the Lorentizan inner product. The Lorentz model is an upper hyper-surface in an $(n+1)$ dimensional Minkowski space with origin point $(\sqrt{-1/c},\;0,\;\cdots,\;0)$.
Special relativity provides a physical interpretation of the Lorentz model by associating the last $n$ components $\mathbf{x}_s$ with spatial dimensions and the 0-th component $x_t$ with the time dimension.
\end{definition}
\begin{definition}[\bf Tangent space]
Given $\mathbf{x}\in\mathbb{L}^{n,c}$, the orthogonal space of $\mathbf{x}$ with respect to the Lorentzian inner product is the tangent space $\mathcal{T}_\mathbf{x}\mathbb{L}^{n,c}$ at point $\mathbf{x}$.
\begin{equation}
    \mathcal{T}_\mathbf{x}\mathbb{L}^{n,c} = \{\mathbf{y}\in\mathbb{R}^{n+1}\;|\;\langle\mathbf{y},\;\mathbf{x}\rangle_\mathcal{L}=0\}.
\end{equation}
\end{definition}
\begin{definition}[\bf Exponential and logarithmic maps]
The exponential map $\text{exp}_\mathbf{x}^c(\cdot):\mathcal{T}_\mathbf{x}\mathbb{L}^{n,c}\rightarrow\mathbb{L}^{n,c}$ is a function that maps any tangent vector from the tangent space at point $\mathbf{x}$ onto the hyperbolic manifold. The exponential map can be formulated as follow:
\begin{equation}
    \text{exp}_\mathbf{x}^c(\mathbf{y}) =  \text{cosh}\left(\sqrt{|c|}\|\mathbf{y}\|_{\mathcal{L}}\right)\mathbf{x}+\frac{\text{sinh}\left(\sqrt{|c|}\|\mathbf{y}\|_{\mathcal{L}}\right)}{\sqrt{|c|}\|\mathbf{y}\|_{\mathcal{L}}}\mathbf{y}.
\end{equation}

The logarithmic map $\text{log}_\mathbf{x}^c(\cdot):\mathbb{L}^{n,c}\rightarrow\mathcal{T}_\mathbf{x}\mathbb{L}^{n,c}$ plays an opposite role, and it can be formulated as follow:
\begin{equation}
\text{log}_\mathbf{x}^c(\mathbf{z})=\frac{\text{cosh}^{-1}\left(c\langle\textbf{x}, \textbf{z}\rangle_\mathcal{L}\right)}{\text{sinh}(\text{cosh}^{-1}\left(c\langle\textbf{x}, \textbf{z}\rangle_\mathcal{L}\right))}(\mathbf{z}-c\langle\mathbf{x}, \mathbf{z}\rangle_\mathcal{L}\;\mathbf{x}).
\end{equation}   
\end{definition}

Hypformer~\cite{yang2024hypformer} introduced Hyperbolic Transformation with Curvatures (HT) and Hyperbolic Readjustment and Refinement with Curvatures (HR). HT enables the curvature to be learnable while preserving relative ordering and remaining disentangled from the normalization term, whereas HR allows the incorporation of several fundamental operations beyond linear transformation, such as dropout, activation functions, and layer normalization.

\begin{definition}[\bf Hyperbolic transformation]
\label{def:HT}
Given a point $\mathbf{x}\in\mathbb{L}^{d,c_1}$ (implies $\mathbf{x}\in\mathbb{R}^{d+1}$), the hyperbolic transformation is defined as following equations:
\begin{align}
    &\text{HT}(\mathbf{x}; f_t, W, c_1, c_2) := (\alpha,\;\beta)^{T},\\
    &\alpha = \sqrt{\frac{c_1}{c_2}\|f_t(\mathbf{x}; W)\|^2_2-\frac{1}{c_2}}\tag*{(time dimension)},\\
    &\beta = \sqrt{\frac{c_1}{c_2}}f_t(\mathbf{x}; W),\tag*{(spatial dimension)}
\end{align}
where $f_t(\mathbf{x};\;W)=W^T\mathbf{x}+b$ denotes the linear transformation with transformation matrix $W\in\mathbb{R}^{(d+1)\times d'}$ and bias $b$. Note that $c_1$ and $c_2$ denote the curvatures before and after the transformation.
\end{definition}

\begin{definition}[\bf Hyperbolic Readjustment and Refinement]
Given a point $\mathbf{x}\in\mathbb{L}^{d,c_1}$, the hyperbolic readjustment and refinement are defined as following equations:
\begin{align}
    &\text{HR}(\mathbf{x}; f_s, W, c_1, c_2) := (\alpha,\;\beta)^{T},\\
    &\alpha = \sqrt{\frac{c_1}{c_2}\|f_s(\mathbf{x}_{[1:]})\|^2_2-\frac{1}{c_2}},\tag*{(time dimension)}\\
    &\beta = \sqrt{\frac{c_1}{c_2}}f_s(\mathbf{x}_{[1:]}),\tag*{(spatial dimension)}
\end{align}
where $f_s(\cdot)$ denotes a function applied to the spatial dimensions, such as dropout, activation function, and LayerNorm.
\end{definition}

\begin{figure*}[t!]
\centering
    \includegraphics[width=0.8\linewidth]{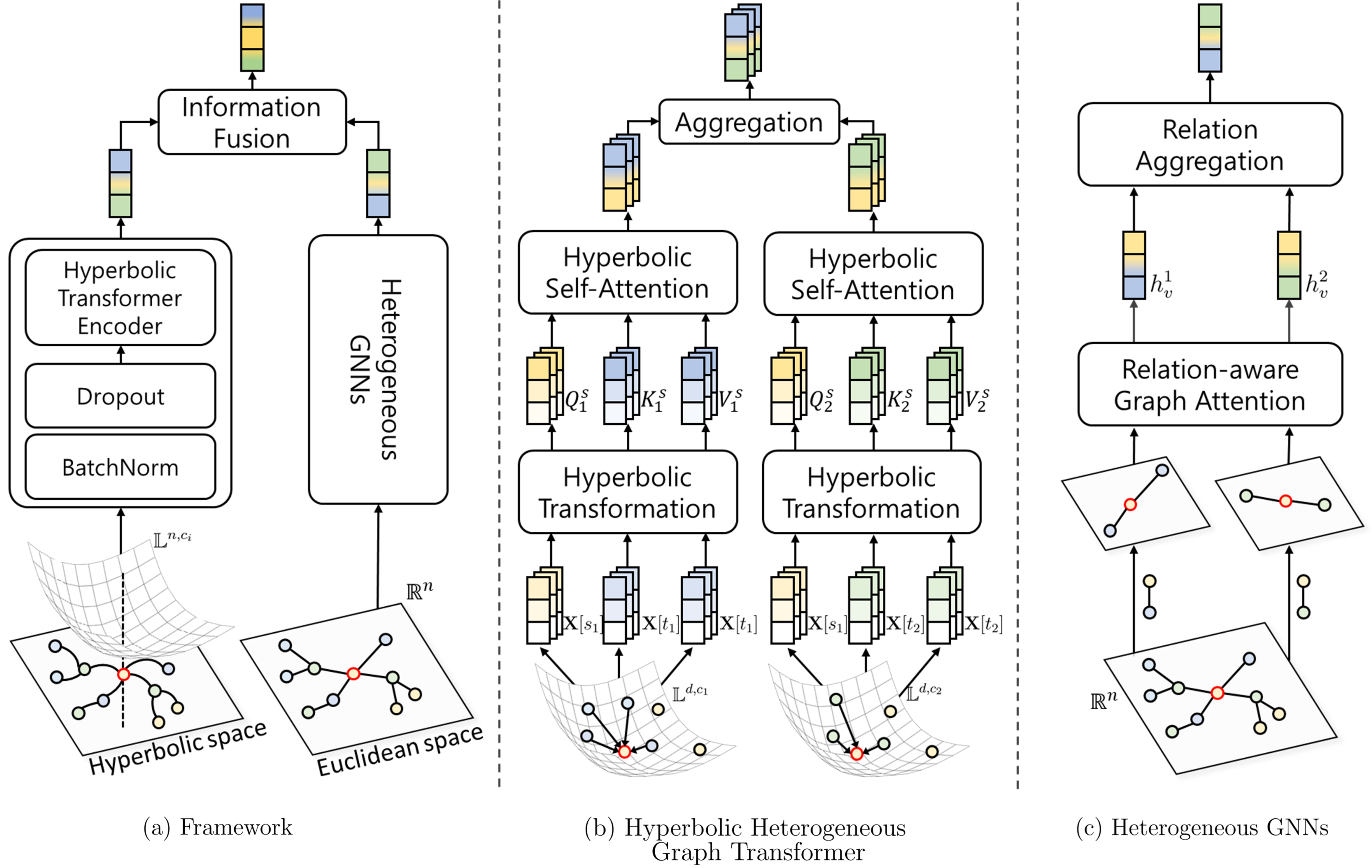}
    \caption{(a) The overall framework of the proposed \algname{}, which consists of two main components: (b) Hyperbolic Heterogeneous Graph Transformer and (c) Heterogeneous GNNs.}
    \label{figure:framework}
\end{figure*}

\section{Methodology}
\label{sec:Methodology}
In this section, we provide a detailed introduction to the proposed \algname{}.
\subsection{Overview}
Figure~\ref{figure:framework} illustrates the overall process of \algname{}. It consists of four main steps as follow:
\begin{enumerate}
    \item First, a relation-aware hyperbolic heterogeneous attention mechanism is proposed to effectively capture the hierarchical structures corresponding to each relation type. Note that, our proposed attention mechanism has linear time complexity.
    \item Second, the representations learned for each relation type through the relation-aware hyperbolic heterogeneous attention mechanism are aggregated to obtain embeddings that captures global structural information.
    \item Third, the heterogeneous GNN layers are used to learn local neighbor information according to each relation type. The relation-specific neighbor representations are aggregated to obtain embeddings that represent the local neighbor information.
    \item Fourth, the final node embeddings are obtained by combining the local neighbor information with the global structural information through the information fusion module.
\end{enumerate}
\subsection{Hyperbolic Heterogeneous Graph Transformer}
\subsubsection{Exponential map to an input Lorentz model}
The first step is to map Euclidean node features $x\in\mathbb{R}^{n}$ to an input Lorentz model $\mathbb{L}^{n,c_i}$ via Exponential map $\text{exp}_\mathbf{o}^{c{_i}}(x):\mathcal{T}_\mathbf{o}\mathbb{L}^{n,c_i}\rightarrow\mathbb{L}^{n,c_i}$, where $n$ is a dimension of node features and $c_i$ is a curvature of $\mathbb{L}^{n,c_i}$. To adopt exponential map, we assume $x$ in the tangent space $\mathcal{T}_{o}\mathbb{L}^{n,c_i}$ at point $o:=0$. We denote the projection of $x$ of into $\mathbb{L}^{n,c_i}$ as $\mathbf{x}\in\mathbb{L}^{n,c_i}$.

% \subsubsection{Positional encoding}
% To incorporate structural position information into hyperbolic representations before hyperbolic attention mechanism, we introduce a relation-specific hyperbolic positional encoding.
% Specifically, the node features $\mathbf{X[s_\tau\|t_{\tau(t)}]}$ is first projected into a relation-specific hyperbolic space using a hyperbolic linear transformation:
% \begin{align}
%     P_\epsilon &= \text{HT}(\mathbf{X[s_{\tau(t)}\|t_{\tau(t)}]},\; f_t,\;W_\epsilon^p,\;c_i;c_\epsilon).
% \end{align}
% The positional embeddings $p_i\in P_\epsilon$ is then scaled by a factor $\gamma$ and added with $\mathbf{x_i}\in\mathbf{X[s_{\tau(t)}\|t_\tau]}$ through the Lorentz midpoint operation as follows:
% \begin{align}
%     \hat{\mathbf{x}}_i^\epsilon = \frac{\mathbf{x_i+\gamma\cdot p_i}}{\sqrt{|c_\epsilon||\mathbf{x_i+\gamma\cdot p_i}||_\mathcal{L}|}},
% \end{align}
% where, $\gamma$ is a scaling factor that controls the influence of the positional embedding on the hyperbolic representation.

\subsubsection{Relation-aware hyperbolic attention mechanism}
For given $\mathbf{X}\in\mathbb{L}^{n,c_i}$ which denotes a set of node features $\mathbf{x}$, we perform hyperbolic attention mechanisms between source and target nodes, where each attention mechanism is conducted within the Lorentz model corresponding to its relation type between them. Specifically, each attention mechanism is distinguished by the source node, target node, and relation type, enabling relation-specific representation learning in the hyperbolic space.
We denote specific relation between a source node and a target node as $(s_{\tau(s)}, r_\epsilon, t_{\tau(t)})$, where $s_{\tau(s)}$ and $t_{\tau(t)}$ denote source and target nodes with node type $\tau(s)$ and $\tau(t)$, respectively, and $r_\epsilon$ indicates relation between them with relation type $\epsilon$. 

Given a specific relation $(s_{\tau(s)}, r_\epsilon, t_{\tau(t)})$, and $\mathbf{X}$, we first perform dropout and BatchNorm operations via HR function as follow equations:
\begin{align}
    \mathbf{X} &= \text{HR}(\mathbf{X},\;f^\epsilon_{BatchNorm}), \\
    \mathbf{X} &= \text{HR}(\mathbf{X},\;f^\epsilon_{dropout}),
\end{align}
where $f_{BatchNorm}^\epsilon$ and $f_{dropout}^\epsilon$ denote relation-specific batch normalization and dropout function, respectively.
Then, with transformation matrices $W^Q_\epsilon, \;W^K_\epsilon$, and $W^V_\epsilon \in \mathbb{R}^{(n+1)\times d}$, we first transform it to $Q_\epsilon, K_\epsilon,\;\text{and}\; V_\epsilon$. Here, $d$ denotes the dimension of latent representations. This process can be formulated as below equations:
\begin{align}
    Q_\epsilon &= \text{HT}(\textbf{X}[s_{\tau(t)}]; f_t, W^Q_\epsilon, c_i, c_\epsilon),\\
    K_\epsilon &= \text{HT}(\textbf{X}[t_{\tau(t)}]; f_t, W^K_\epsilon, c_i, c_\epsilon),\\
    V_\epsilon &= \text{HT}(\textbf{X}[t_{\tau(t)}]; f_t, W^V_\epsilon, c_i, c_\epsilon),
\end{align}
where $Q_\epsilon,\;K_\epsilon$, and $\;V_\epsilon\in \mathbb{L}^{d,c_\epsilon}$. Here, $c_\epsilon$ is the learnable curvature of relation-specific hyperbolic space $\mathbb{L}^{d,c_\epsilon}$. 

During the computation of attention output and weighted-sum aggregation, the time dimension in the Lorentz model is not included in the process. Therefore, only the spatial dimensions are used for these operations. 
After slicing the values along the spatial dimension, the ReLU-based kernel function is applied to ensure non-negativity and enable the decomposition of the softmax function into a linear form. This function transforms features and allows the attention mechanism to be computed in linear time, while maintaining the kernel function properties of the original softmax-based attention mechanism. 
This process can be formulated as follows:
\begin{align}
    \phi(\mathbf{X}) &= \frac{\text{ReLU}(\mathbf{X}_{[1:]})+\alpha}{\|\beta\|},\\ 
    Q^s_\epsilon, K^s_\epsilon, V^s_\epsilon  &= \phi(Q_\epsilon), \phi(K_\epsilon), \phi(V_\epsilon),
\end{align}
where $\phi(\cdot)$ is a kernel function, $\alpha$ is a small positive constant that is added to prevent numerical instability when the feature values or their norms approach zero, and $\beta$ is a learnable normalization parameter used to stabilize feature scaling after the transformation. This design prevents the kernel values from exploding or vanishing during the attention mechanism, especially in the hyperbolic space, where feature magnitudes are highly sensitive to curvature.

After feature transformation with kernel function, the relation-specific attention output is computed as follows:
\begin{align}
\label{eq:12}
    H^s_\epsilon =\frac{Q^s_\epsilon (K^{^sT}_\epsilon V^s_\epsilon)}{Q^s_\epsilon (K^{^sT}_\epsilon\;\textbf{1})},
\end{align}
where \textbf{1} is an all-ones vector and $H^s_\epsilon\in\mathbb{L}^{d,c_\epsilon}$ is the relation-specific attention output for relation type $\epsilon$.

Given the attention output $H^s_\epsilon$, the separated time dimension is concatenated back with $H^s_\epsilon$ (i.e., the spatial dimension) to reconstruct the full Lorentz representation, as follows:
\begin{align}
    H^t_\epsilon &= \sqrt{\|H^s_\epsilon\|^2_2-1/c_\epsilon},\\
    H_\epsilon &= \text{concat}(H^t_\epsilon\;,H^s_\epsilon),
\end{align}
where $H^t_\epsilon$ represents the time dimension component reconstructed to satisfy the Lorentz manifold constraint. By concatenating the reconstructed time-dimension component with the spatial-dimension component (i.e., $H^s_\epsilon$), we obtain the Lorentz representations $H_\epsilon$, ensuring that the relation-specific embeddings lie on the relation-specific hyperbolic space $\mathbb{L}^{d,c_\epsilon}$.

\subsubsection{Relation aggregation}
After obtaining the relation-specific representations $H_\epsilon\in\mathbb{L}^{d,c_\epsilon}$ from each relation-specific hyperbolic space, we first transform them into a unified output hyperbolic space to ensure geometric consistency across different relations types.
This transform is performed through HT function defined for each relation type as follow:
\begin{align}
    H'_\epsilon = \text{HT}(H_\epsilon;f_t,W_o,c_\epsilon,c_o)\; (\forall\epsilon\in T_\mathcal{E}),
\end{align}
where $H'_\epsilon\in\mathbb{L}^{d,c_o}$ are transformed relation-specific representations, $W_o\in\mathbb{R}^{(d+1)\times d}$ is the transformation matrix, and $c_o$ is the curvature of the output Lorentz model $\mathbb{L}^{d,c_o}$.
This transformation aligns all relation-specific representations within a unified Lorentz model, making them compatible with aggregation.

After the transformation, the representations from all relation types are aggregated through a mean operation to obtain the unified representation as follow:
\begin{align}
    H_\mathcal{T} = \frac{1}{|T_\mathcal{E}|}\sum_{\epsilon\in T_\mathcal{E}}\text{log}^{c_o}_\mathbf{o}(H'_\epsilon),
\end{align}
where $H_\mathcal{T}\in\mathbb{R}^{|\mathcal{V}|\times d}$ denotes output representations from hyperbolic transformer, $T_\epsilon$ represents a set of relation types, and $\text{log}^{c_o}_\mathbf{o}(\cdot):\mathbb{L}^{d,c_o}\rightarrow\mathcal{T_\mathbf{o}}\mathbb{L}^{d,c_o}$ indicates logarithmic map, which projects $H'_\epsilon$ from $\mathbb{L}^{d,c_o}$ onto its tangent space $\mathcal{T}_\mathbf{o}\mathbb{L}^{d,c_o}$ at the point $\mathbf{o}:= 0$.

\subsubsection{Multi-head attention mechanism}
We introduce multi-head attention in the hyperbolic space to enhance representations from hyperbolic heterogeneous graph transformer. Specifically, we divide attention mechanisms into $K$ independent attention mechanisms, conduct them in parallel, and then concatenate the representations from each attention mechanisms to obtain final representation $H_\tau\in\mathbb{R}^{|\mathcal{V}|\times d}$. Here, $\mathcal{V}$ denotes a set of nodes. This process can be formulated as:
\begin{align}
    H_\tau=\;\|^{K}_{k=1}\;\left(\frac{1}{|T_\mathcal{E}|}\sum_{\epsilon\in T_\mathcal{E}}\text{log}^{c_o}_\mathbf{o}(H^{'k}_\epsilon)\right).
\end{align}

\begin{algorithm}[t!]
\caption{\small{Hyperbolic heterogeneous graph transformer}}
\label{alg:process}
\begin{algorithmic}
\begin{small}
\REQUIRE The set of node features $X$, The set of relation types $T_\mathcal{E}$, The number of attention heads $K$.
%\ENSURE Prediction for downstream tasks.
\end{small}
\end{algorithmic}
\begin{algorithmic}[1]
\begin{small}
\setstretch{1.2}
\STATE{\textbf{(1) Exponential map to input Lorentz model}}
\STATE{$\mathbf{X}\leftarrow \text{exp}_\mathbf{o}^{c_i}(X)$;}
\FOR{$k=1,\cdots, K$}
\STATE{\textbf{(2) Relation-aware hyperbolic attention mechanism}}
\FOR{$\epsilon\in T_\mathcal{E}$}
\STATE{$\mathbf{X}\leftarrow\text{HR}(\mathbf{X}, f_{BatchNorm}^\epsilon)$;}
\STATE{$\mathbf{X}\leftarrow\text{HR}(\mathbf{X}, f_{dropout}^\epsilon)$;}
\STATE{$Q_\epsilon\leftarrow\text{HT}(\textbf{X}[s_{\tau(t)}]; f_t, W^Q_\epsilon, c_i, c_\epsilon)$;}
\STATE{$K_\epsilon\leftarrow\text{HT}(\textbf{X}[t_{\tau(t)}]; f_t, W^K_\epsilon, c_i, c_\epsilon)$;}
\STATE{$V_\epsilon\leftarrow\text{HT}(\textbf{X}[t_{\tau(t)}]; f_t, W^V_\epsilon, c_i, c_\epsilon)$;}
\STATE{$Q^s_\epsilon, K^s_\epsilon, V^s_\epsilon \leftarrow \phi(Q_\epsilon), \phi(K_\epsilon), \phi(V_\epsilon)$;}
\STATE{$H^s_\epsilon \leftarrow \frac{Q^s_\epsilon (K^{^sT}_\epsilon V^s_\epsilon)}{Q^s_\epsilon (K^{^sT}_\epsilon\;\textbf{1})}$;}
\STATE{$H^t_\epsilon \leftarrow \sqrt{\|H^s_\epsilon\|^2_2-1/c_\epsilon}$;}
\STATE{$H_\epsilon \leftarrow \text{concat}(H^t_\epsilon\;,H^s_\epsilon)$;}
\STATE{$H^{'k}_\epsilon \leftarrow \text{HT}(H_\epsilon;f_t,W_o,c_\epsilon,c_o)$;}
\ENDFOR
\ENDFOR
\STATE{\textbf{(3) Relation-aggregation and Multi-head attention}}
\STATE{$H_\mathcal{T}\leftarrow\|^{K}_{k=1}\frac{1}{|T_\mathcal{E}|}\sum_{\epsilon\in T_\mathcal{E}}\text{log}^{c_o}_\mathbf{o}(H'^{k}_\epsilon);$}
\RETURN{$H_\mathcal{T}$;}
\end{small}
\end{algorithmic}
\end{algorithm}

\subsection{Heterogeneous GNNs}
\subsubsection{Heterogeneous graph attention mechanism}
While hyperbolic heterogeneous graph transformer effectively captures global structural information across different relation types, it is also essential to aggregate local neighbor information to enhance the semantic completeness of node representations. In heterogeneous graphs, local neighbors often contain rich type-dependent information that inherent relational semantics, which may not be fully addressed in hyperbolic heterogeneous graph transformer.

To address this, we use heterogeneous GNNs that perform localized message-passing among different types of neighbors. For a given embedding target node $v$ and its neighbors $\mathcal{N}_\epsilon(v)$ connected through the relation type $\epsilon$, the attention score $\alpha^\epsilon_{v,u}$ between $v$ and the node $u\in\mathcal{N}_\epsilon(v)$ is computed as:

\begin{align}
    \alpha^\epsilon_{v,u} = \frac{exp(\text{a}_\epsilon^\top\text{LeakyReLU}(W_\epsilon\cdot[h_v||h_u]))}{\sum_{k\in\mathcal{N}(v)}exp(\text{a}_\epsilon^\top\text{LeakyReLU}(W_\epsilon\cdot[h_v||h_k]))}
\end{align}

where $h_v,\; h_u\in \mathbb{R}^n$ is denotes features of node $v$ and $u$, respectively, $W_\epsilon\in\mathbb{R}^{d\times 2n}$ denotes relation-specific transformation matrix and $\mathbf{a}_\epsilon$ is a learnable attention vector used to compute weights for relation type $\epsilon$.

After obtaining attention scores, the relation-specific embedding of the target node $v$ for relation type $\epsilon$ is obtained by aggregating the neighbor features weighted by these attention scores:
\begin{align}
    h_v^\epsilon = \sigma\left(\sum_{u\in\mathcal{N}(v)}\alpha^\epsilon_{v,u}W_\epsilon h_u\right),
\end{align}
where $\sigma$ denotes a activation function and $h^\epsilon_v\in\mathbb{R}^d$ denotes the relation-specific embedding.
\subsubsection{Relation aggregation}
Finally, to obtain a unified node representation that aggregates information from all relation types, the relation-specific embeddings are aggregated through a mean operation:
\begin{align}
    h'_v = \frac{1}{|T_\mathcal{E}|}\sum_{\epsilon\in T_\mathcal{E}}h_v^\epsilon,
\end{align}
where $h'_v\in\mathbb{R}^d$ denotes embeddings of node $v$ obtained from Heterogeneous GNNs. This aggregation enables \algname{} to learn type-dependent neighbor information while preserving the heterogeneity of relations.

Furthermore, heterogeneous GNNs performs a multi-head attention mechanism, which can be formulated as follows.

\begin{align}
    h'_v=\;\|^{K}_{k=1}\;\left(\frac{1}{|T_\mathcal{E}|}\sum_{\epsilon\in T_\mathcal{E}}h_v^{\epsilon,k}\right).
\end{align}

\begin{algorithm}[t!]
\caption{\small{Heterogeneous GNNs and Information Fusion}}
\label{alg:process}
\begin{algorithmic}
\begin{small}
\REQUIRE Embedding target node $v$, The set of neighbor nodes of $v$ $\mathcal{N}(v)$, The set of relation types $T_\mathcal{E}$, The number of attention heads $K$, lambda $\lambda$.
%\ENSURE Prediction for downstream tasks.
\end{small}
\end{algorithmic}
\begin{algorithmic}[1]
\begin{small}
\setstretch{1.2}
\STATE{\textbf{(1) Heterogeneous graph attention}}
\FOR{$\epsilon\in T_\mathcal{E}$}
\FOR{$k=1,\cdots,K$}
\FOR{$u\in\mathcal{N}(v)$}
\STATE{$a^\epsilon_{v,u}\leftarrow\mathbf{a}_\epsilon^\top\text{LeakyReLU}(W_\epsilon\cdot[h_v\|h_u]);$}
\STATE{$\alpha^\epsilon_{v,u}\leftarrow\text{softmax}(exp(a^\epsilon_{v,u}))$}
\ENDFOR
\STATE{$h_v^\epsilon \leftarrow \sigma\left(\sum_{u\in\mathcal{N}(v)}\alpha^\epsilon_{v,u}W_\epsilon h_u\right)$}
\ENDFOR
\STATE{\textbf{(2) Relation aggregation}}
\STATE{$h'^{k}_v \leftarrow \frac{1}{|T_\mathcal{E}|}\sum_{\epsilon\in T_\mathcal{E}}h_v^\epsilon$}
\ENDFOR
\STATE{$h'_v\leftarrow\;\|^{K}_{k=1}\;h'^{k}_v$}
\STATE{\textbf{(3) Information fusion}}
\STATE{$z_v \leftarrow\lambda\cdot(H_\mathcal{T}(v)) + (1-\lambda)\cdot h'_v\;$}
\RETURN{$z_v$;}
\end{small}
\end{algorithmic}
\end{algorithm}

\subsection{Information Fusion}
To obtain final node representations, we integrate the global structural information learned from the hyperbolic heterogeneous graph transformer and the local neighbor information learned from the heterogeneous GNNs.
This information fusion allows \algname{} to capture both global hierarchical semantics and local relationships within heterogeneous graphs. For given a embedding target node $v$, the corresponding final node embedding $z_v$ is defined through this information fusion as follows:
\begin{align}
    z_v = \lambda\cdot(H_\mathcal{T}(v)) + (1-\lambda)\cdot h'_v\;,
\end{align}
where, $H_\mathcal{T}(v)\in H_\mathcal{T}$ and $h'_v$ represent the embeddings of node $v$ which obtained from hyperbolic heterogeneous graph transformer and heterogeneous GNNs, respectively, and $\lambda\in[0,1]$ is a hyperparameter that adjust magnitude of global and local information. In this paper, we set $\lambda$ to 0.5.
\subsection{Model Training}
We use the following linear transformation with non-linear activation function $f(\cdot)$ to map node embeddings $z_v\in\mathbb{R}^d$ into a vector space with the desired output dimension, conducting node classification task:
\begin{align}
    f(z_v) = \sigma(W_o\cdot z_v),
\end{align}
where $W_o\in\mathbb{R}^{d_o\times d}$ denotes transformation matrix, $d_o$ denotes the dimension of output vector, and $\sigma$ is the non-linear activation function.

For node classification, we trained \algname{} by minimizing cross-entropy loss $\mathcal{L}$ expressed as 
\begin{align}
    \mathcal{L} &=-\sum_{v\in V_t}\sum_{c=1}^Cy_v[c]\cdot\text{log}\left(f(z_v)[c]\right),
\end{align}
where $v_t$ is the target node set extracted from the labeled node set, $C$ is the number of classes, $y_v$ is the one-hot encoded label vector for node $v$, and $f(z_v)$ is a vector predicting the label probabilities of node $v$.

\subsection{Time Complexity}
\label{sec:time_complex}
\algname{} consists of two main components for heterogeneous graph representation learning: hyperbolic heterogeneous graph transformer and heterogeneous GNNs. As shown in Equation~\ref{eq:12}, the hyperbolic heterogeneous graph transformer performs an inner product between $K_\epsilon^{sT}$ and $V_\epsilon^s$ within the Lorentz model corresponding to each relation type $\epsilon$. This process has a time complexity of $O(|T_\mathcal{E}|\cdot d^2\cdot|\mathcal{T}|)$, where $T_\mathcal{E}$ and $\mathcal{T}$ denotes a set of relation types and target nodes, respectively. Subsequently, an inner product is performed between $Q_\epsilon^s$ and the output of $K^{^sT}_\epsilon V^s_\epsilon$, which has a time complexity of $O(|T_\mathcal{E}|\cdot d^2\cdot(|\mathcal{S}+\mathcal{T}|))$, where $\mathcal{S}$ denotes set of source nodes. Therefore, the total time complexity of hyperbolic heterogeneous graph transformer is $O(|\mathcal{S}+\mathcal{T}|)\approx O(N)$. Meanwhile, the time complexity of heterogeneous GNNs is $O(|T_\mathcal{E}|\cdot(N+E))$. Consequently, our proposed model \algname{} has a linear-time complexity overall, which enables efficient and scalable heterogeneous graph representation learning even on large-scale graphs.

\section{Experimental Evaluations}
In this section, we analyze the efficiency of our proposed model \algname{}, through with four real-world heterogeneous graph datasets and synthetic heterogeneous graphs. We compare \algname{} with several state-of-the-art heterogeneous graph embedding models. Our experiments are conducted to answer the following research questions (RQs).
\begin{itemize}
    \item {\bf RQ1 :} Does \algname{} outperform the state-of-the-art methods in node classification task?
    \item {\bf RQ2 :} How does each main component of \algname{} contribute to the learning of heterogeneous graph representations?
    \item {\bf RQ3 :} How do multiple hyperbolic spaces in \algname{} affect heterogeneous graph representation learning?
    \item {\bf RQ4 :} Is \algname{} more efficient than existing heterogeneous graph embedding models?
\end{itemize}
\label{sec:Experiments}

\subsection{Datasets}
To evaluate the performance of \algname{} on node classification task, we use three-real world heterogeneous graph datasets. Table~\ref{tab:datastats} shows the statistics of these datasets. Additionally, detailed description of the datasets are provided as below:
\begin{itemize}
    \item {\textbf{IMDB}\footnote{https://www.imdb.com/}} is an online database pertaining to movies and television programs. It comprises three types of nodes \{Movie (M), Director (D), Actor (A)\} and two types of links \{MD, MA\}. The movie type nodes are labeled into three classes based on the movie's genre \{Action, Drama, Comedy\}. The features of nodes are represented as bag-of-words of keywords.
    \item {\textbf{DBLP}\footnote{https://dblp.uni-trier.de/}} is a citation network that comprises three types of nodes \{Author (A), Paper (P), Conference (C)\} and two types of links \{AP, PC\}. The author type nodes are labeled into four classes based on the author's research area \{Database, Data Mining, Machine Learning, Information Retrieval\}. The features of nodes are represented as bag-of-words of keywords.
    \item {\textbf{ACM}\footnote{https://dl.acm.org/}} is a citation network that comprises three types of nodes \{Paper (P), Author (A), Subject (S)\} and two types of links \{PA, PS\}. The paper type nodes are labeled into three classes based on the paper's subject area \{Database, Wireless Communication, Data Mining\}. The features of nodes are represented as bag-of-words of plots.
\end{itemize}

\begin{table}[t]
\centering
\caption{Statistics of real-world datasets}
\label{tab:datastats}
\resizebox{\columnwidth}{!}{%
\begin{tabular}{ccccc} \hline
Dataset & \# Nodes & \# Links & \# Classes & \# Features \\\hline
IMDB    & \begin{tabular}[c]{@{}c@{}}Movie (M) : 4,661\\ Director (D) : 2,270\\ Actor (A) : 5,841\end{tabular}   & \begin{tabular}[c]{@{}c@{}}M-D : 4,661\\ M-A : 13,983\end{tabular} & 3  & 1,256\\\hline
DBLP    & \begin{tabular}[c]{@{}c@{}}Author (A) : 4,057\\Paper (P) : 14,328\\ Conference (C) : 20\end{tabular}   & \begin{tabular}[c]{@{}c@{}}A-P : 19,645\\ P-C : 14,328\end{tabular} & 4  & 334\\\hline
ACM    & \begin{tabular}[c]{@{}c@{}}Paper (P) : 3,020\\ Author (A) : 5,912\\ Subject (S) : 57\end{tabular}   & \begin{tabular}[c]{@{}c@{}}P-A : 9,936\\ P-S : 3,025\\\end{tabular} & 3 & 1,902\\\hline
\end{tabular}%
}
\end{table}

\begin{table*}[]
\centering
\caption{Experimental results (\%) for the node classification task. The best and second-best performers are bolded and underlined, respectively.}
\label{tab:node_classification}
\resizebox{\textwidth}{!}{%
\begin{tabular}{ccccccccccccccc}
\hline
\multirow{2}{*}{Dataset} &
  \multirow{2}{*}{Metric} &
  \multicolumn{2}{c}{\lowercase\expandafter{\romannumeral1}} &
  \multicolumn{2}{c}{\lowercase\expandafter{\romannumeral2}} &
  \multicolumn{5}{c}{\lowercase\expandafter{\romannumeral3}} &
  \multicolumn{4}{c}{\lowercase\expandafter{\romannumeral4}} \\ 
\cmidrule(lr){3-4} \cmidrule(lr){5-6} \cmidrule(lr){7-11} \cmidrule(lr){12-15}
 &
  &
  GCN &
  GAT &
  HGCN &
  Hypformer &
  HAN &
  MAGNN &
  GTN &
  HGT &
  Simple-HGN &
  SHAN &
  HHGAT &
  MSGAT &
  \algname{} \\ 
\hline

\multirow{2}{*}{IMDB} &
  Macro-F1 &
  54.38 & 57.06 & 57.98 & 65.12 &
  59.26 & 61.13 & 62.73 & 62.87 & 67.02 &
  66.75 & 66.38 & \underline{68.91} & \bf70.47 \\
 &
  Micro-F1 &
  54.89 & 57.53 & 58.68 & 68.15 &
  60.10 & 61.89 & 64.26 & 63.29 & 69.31 &
  69.99 & 70.28 & \underline{70.45} & \bf72.56 \\

\hline

\multirow{2}{*}{DBLP} &
  Macro-F1 &
  89.53 & 91.29 & 92.70 & 94.18 &
  92.81 & 93.62 & 93.83 & 93.96 & 94.11 &
  94.46 & 93.76 & \underline{94.51} & \bf95.68 \\
 &
  Micro-F1 &
  90.06 & 92.15 & 93.39 & 94.36 &
  93.43 & 94.28 & 94.18 & 94.02 & 94.73 &
  94.98 & 94.56 & \underline{95.28} & \bf96.04 \\

\hline

\multirow{2}{*}{ACM} &
  Macro-F1 &
  88.76 & 88.95 & 90.26 & 93.81 &
  91.26 & 91.29 & 92.54 & 91.79 & 93.40 &
  93.71 & 93.62 & \underline{93.84} & \bf94.24 \\
 &
  Micro-F1 &
  88.53 & 89.06 & 90.71 & 93.90 &
  92.47 & 92.70 & 92.56 & 92.07 & 93.13 &
  \underline{94.32} & 93.14 & 93.95 & \bf94.35 \\

\hline
\end{tabular}%
}
\end{table*}

\subsection{Competitors}
We compare \algname{} with several state-of-the-art GNNs categorized four groups:
\begin{enumerate}[\bf i)]
    \item {\textbf{Euclidean homogeneous models}}: GCN, and GAT.
    \item {\textbf{Hyperbolic homogeneous models}}: HGCN, and Hypformer.
    \item {\textbf{Euclidean heterogeneous models}}: HAN, MAGNN, GTN, HGT, and Simple-HGN. 
    \item {\textbf{Hyperbolic heterogeneous models}}: SHAN, HHGAT, and MSGAT.
\end{enumerate}
For homogeneous models, features are processed to be homogeneous for pair comparison with heterogeneous models. Details of the competitors are provided as follows:
\begin{itemize}
    \item {\textbf{GCN}}~\cite{kips2017iclr} performs graph convolution operations in the Fourier domain for homogeneous graphs.
    \item {\textbf{GAT}}~\cite{velickovic2018iclr} introduces graph attention mechanisms into the graph convolution operation for homogeneous graphs.
    \item {\textbf{HGCN}}~\cite{chami2019hyperbolic} proposes graph neural networks that uses the hyperbolic space as an embedding space to effectively learn complex structures.
    \item {\textbf{Hypformer}}~\cite{yang2024hypformer} proposes a hyperbolic graph transformer architecture in which operations are performed directly within the hyperbolic space, reducing mapping distortions and achieving efficient representation learning with linear-time complexity.
    \item {\textbf{HAN}}~\cite{wang2019heterogeneous} proposes a graph attention network for heterogeneous graphs, which incorporates both node-level and semantic-level attention mechanisms.
    \item {\textbf{MAGNN}}~\cite{fu2020magnn} introduces intra-metapath and inter-metapath aggregation to incorporate intermediate semantic nodes and multiple metapaths, respectively.
    \item {\textbf{GTN}}~\cite{yun2019graph} transforms heterogeneous graphs into multiple metapath graphs through Graph Transformer layers, facilitating more effective node representations.
    \item {\textbf{HGT}}~\cite{hu2020heterogeneous} proposes a heterogeneous subgraph sampling algorithm to model Web-scale graph data and utilize node and link-type dependent parameters to capture heterogeneous relations.
    \item {\textbf{Simple-HGN}}~\cite{lv2021we} extends graph attention mechanism by including link type information into attention score calculation, enabling Simple-HGN to handle multiple type of links in heterogeneous graphs.
    \item {\textbf{SHAN}}~\cite{li2023multi} proposes hyperbolic heterogeneous graph attention networks to learn multi-order relations from simplical complexes sampled from heterogeneous graphs.
    \item {\textbf{HHGAT}}~\cite{park2024hyperbolic} introduces hyperbolic heterogeneous graph attention networks with a single hyperbolic space to learn complex structures within heterogeneous graphs.
    \item {\textbf{MSGAT}}~\cite{park2024multi} proposes multi hyperbolic space-based heterogeneous graph attention networks to learn various complex structures based on metapaths.
\end{itemize}

\subsection{Implementation Details}
For the baselines including \algname{}, we randomly initialize model parameters and use the adamW~\cite{LoshchilovH19} optimizer with a learning rate of 0.0001 and weight decay of 0.0005. We set the dropout rate to 0.5, and the dimensional of final node embedding to 64. For multi-head attention baselines, the number of attention heads for each model was set to 8. For \algname{}, the number of heads in the hyperbolic transformer was set to 2, while the number of heads in the heterogeneous GNNs was set to 8. The baseline models are trained for 300 epochs, and the model with the lowest validation loss is used for testing. For the metapath-based heterogeneous graph embedding models, the metapath settings follow the specifications outlined in their papers. All downstream tasks were conducted ten times, and we report the average values of evaluation metrics.

\subsection{Node Classification (RQ1)}
Node classification was performed by applying support vector machines on embedding vectors of labeled nodes. 

As shown in Table~\ref{tab:node_classification}, \algname{} achieves the best performance across all datasets (i.e., IMDB, DBLP and ACM) in both Macro-F1 and Mirco-F1 scores, outperforming all baselines. These results demonstrate the effectiveness of our proposed model \algname{} in learning both local and global structural information within heterogeneous graphs. We discuss three main empirical findings based on the results.

First, the comparison between \algname{} and Hypformer highlights the importance of learning heterogeneous relational information within heterogeneous graphs. While Hypformer demonstrates strong performance as a hyperbolic transformer model, it is designed for homogeneous graphs and therefore lacks the ability to learn semantic information inherent in multiple relation types within heterogeneous graphs. In contrast, \algname{} extends this transformer architecture to heterogeneous graphs by introducing relation-specific attention mechanism and distinct hyperbolic space for each relation type. This allows \algname{} to effectively learn relation-dependent semantic representations, leading to significant improvements.

Second, the comparison between Euclidean heterogeneous models (i.e., category \romannumeral 3) and hyperbolic heterogeneous models (i.e., category \romannumeral 4) demonstrates distinct ability to learn hierarchical structures within heterogeneous graphs. Since the Euclidean space expands polynomially, Euclidean heterogeneous models such as GTN, HGT struggle to represent hierarchical structures that are observed in heterogeneous graphs. In contrast, hyperbolic heterogeneous models use the hyperbolic space as an embedding space, which expands exponentially and thus enables a more effective representation of hierarchical structures.

Finally, compared to previous hyperbolic heterogeneous models that rely heavily on tangent space, \algname{} performs attention mechanisms entirely within the hyperbolic space. This fully hyperbolic space operations minimizes mapping distortions and enables more stable and efficient heterogeneous graphs representation learning.

\begin{figure*}[t!]
\captionsetup[subfigure]{justification=centering}
\centering
    \begin{minipage}[b]{0.3\linewidth}
        \centering
        \includegraphics[width=\linewidth]{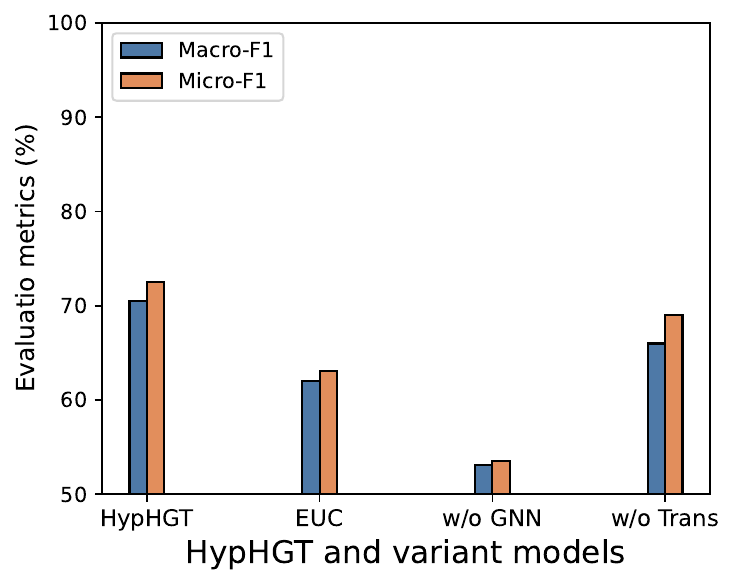}
        \subcaption{Results on IMDB dataset.}
    \end{minipage}
    \begin{minipage}[b]{0.3\linewidth}
        \centering
        \includegraphics[width=\linewidth]{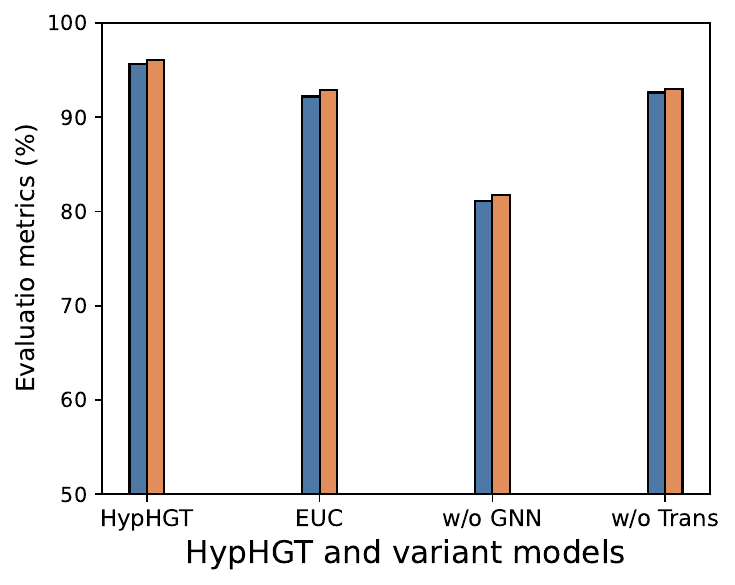}
        \subcaption{Results on DBLP dataset.}
    \end{minipage}
    \begin{minipage}[b]{0.3\linewidth}
        \centering
        \includegraphics[width=\linewidth]{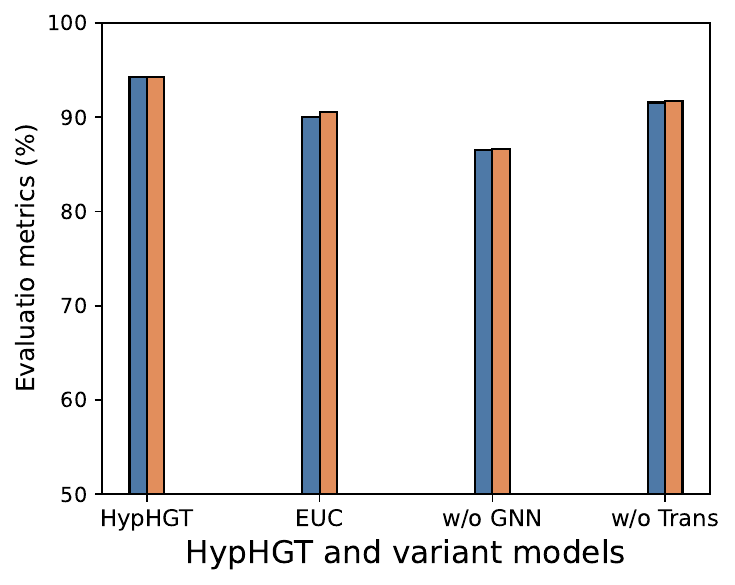}
        \subcaption{Results on ACM dataset.}
    \end{minipage}
\caption{Results of the ablation study.}
\label{fig:ablation}
\end{figure*}

\begin{figure}[t!]
\captionsetup[subfigure]{justification=centering}
\centering
    \begin{minipage}[b]{0.49\columnwidth}
        \centering
        \includegraphics[width=\linewidth]{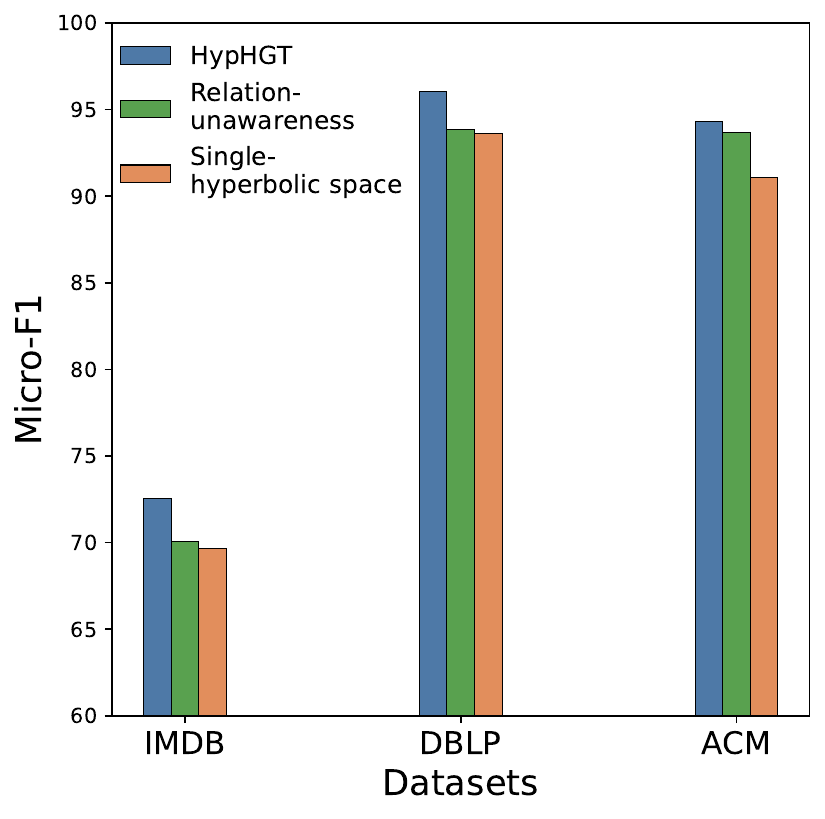}
        \subcaption{Micro-F1}
    \end{minipage}
    \begin{minipage}[b]{0.49\linewidth}
        \centering
        \includegraphics[width=\linewidth]{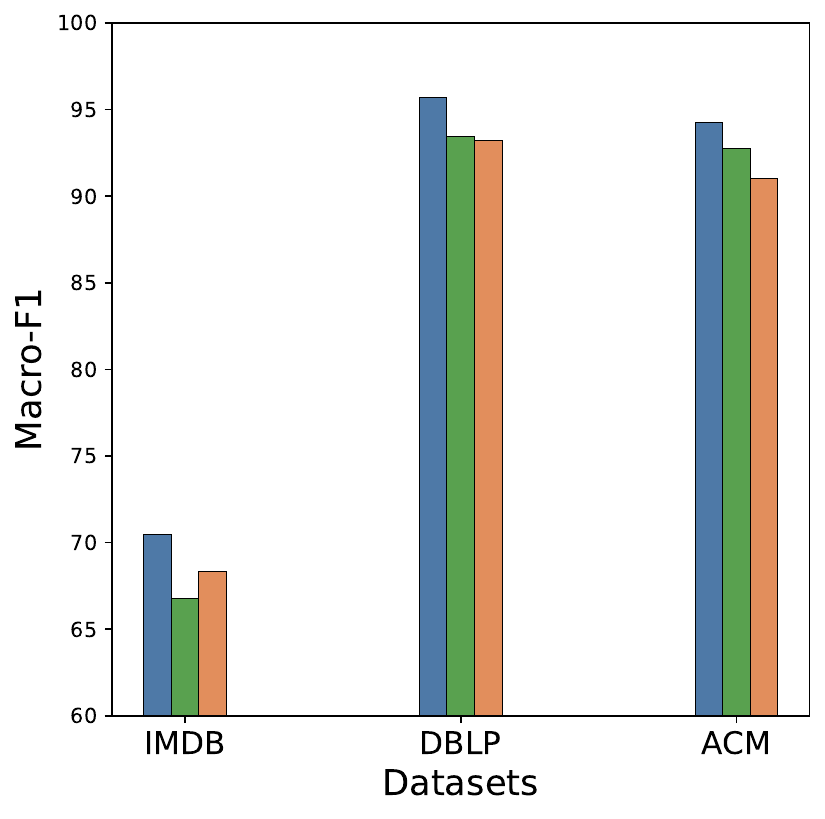}
        \subcaption{Macro-F1}
    \end{minipage}
\caption{Node classification accuracy according to the configuration of hyperbolic spaces.}
\label{figure:space_analysis}
\end{figure}

\subsection{Ablation Study (RQ2)}
We compose three variants of \algname{} to validate the effectiveness of each component of \algname{}. Specifically, 1) \textbf{HypHGT}$_{\textbf{EUC}}$ replaces the hyperbolic space with Euclidean space as the embedding space, 2) \textbf{HypHGT}$_{\textbf{w/o GNN}}$ removes heterogeneous GNN layers, and 3) \textbf{HypHGT}$_{\textbf{w/o Trans}}$ removes hyperbolic heterogeneous graph transformer layers. Figure~\ref{fig:ablation} represents the results of the ablation study. 

First, when comparing \algname{} with \textbf{HypHGT}$_{\textbf{EUC}}$, the results demonstrate that the hyperbolic space is more effective in capturing the hierarchical structures within heterogeneous graphs. Second, the comparison between \algname{} and \textbf{HypHGT}$_{\textbf{w/o GNN}}$ demonstrates that, since graphs are inherently composed of local relations between entities, using the hyperbolic heterogeneous graph transformer alone fails to capture local neighborhood information, which is a critical limitation. Consequently, the \textbf{HypHGT}$_{\textbf{w/o GNN}}$ variant shows the most significant performance decrease. Finally, when comparing \algname{} with \textbf{HypHGT}$_{\textbf{w/o Trans}}$, although \textbf{HypHGT}$_{\textbf{w/o Trans}}$ achieves competitive performance, it still struggles to capture the global hierarchical structures. This limitation arises because the graph convolution operations in GNNs are primarily localized within neighborhoods, making it difficult to capture long-range dependencies in heterogeneous graphs. In contrast, \algname{} overcomes this limitation by integrating the hyperbolic heterogeneous graph transformer with heterogeneous GNNs, enabling the model to capture both global structural information and local neighborhood dependencies effectively.

\begin{figure*}[t!]
\captionsetup[subfigure]{justification=centering}
\centering
    \begin{minipage}[b]{0.33\linewidth}
        \centering
        \includegraphics[width=\linewidth]{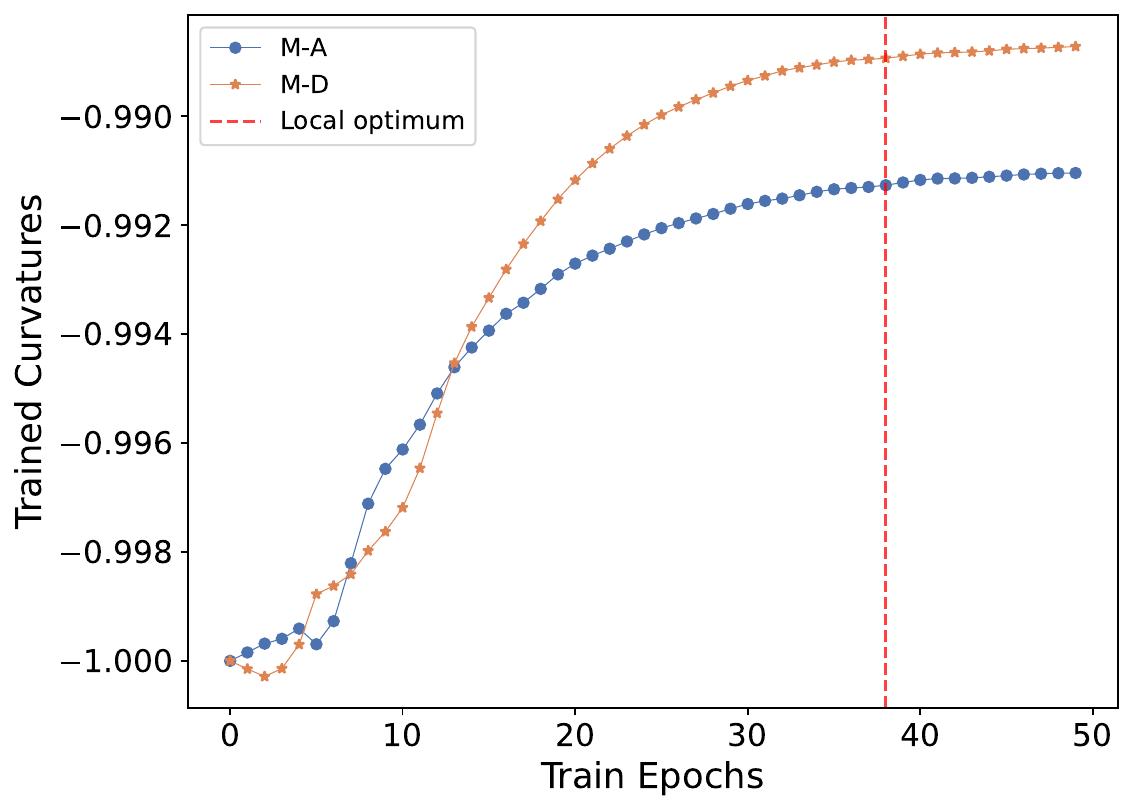}
        \subcaption{Results on IMDB dataset.}
        \vspace{0.25em}
    \end{minipage}
    \begin{minipage}[b]{0.32\linewidth}
        \centering
        \includegraphics[width=\linewidth]{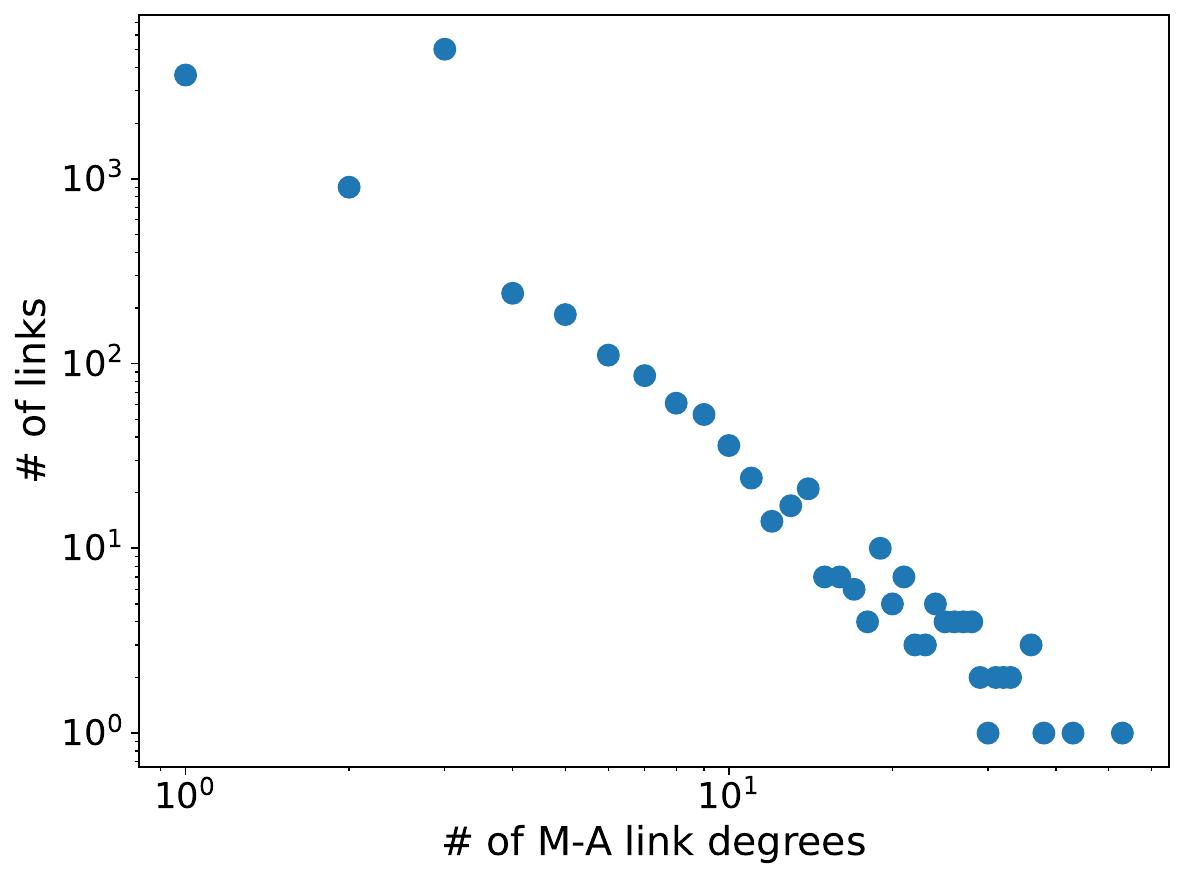}
        \subcaption{Movie-Actor relation.}
        \vspace{0.25em}
    \end{minipage}
    \begin{minipage}[b]{0.32\linewidth}
        \centering
        \includegraphics[width=\linewidth]{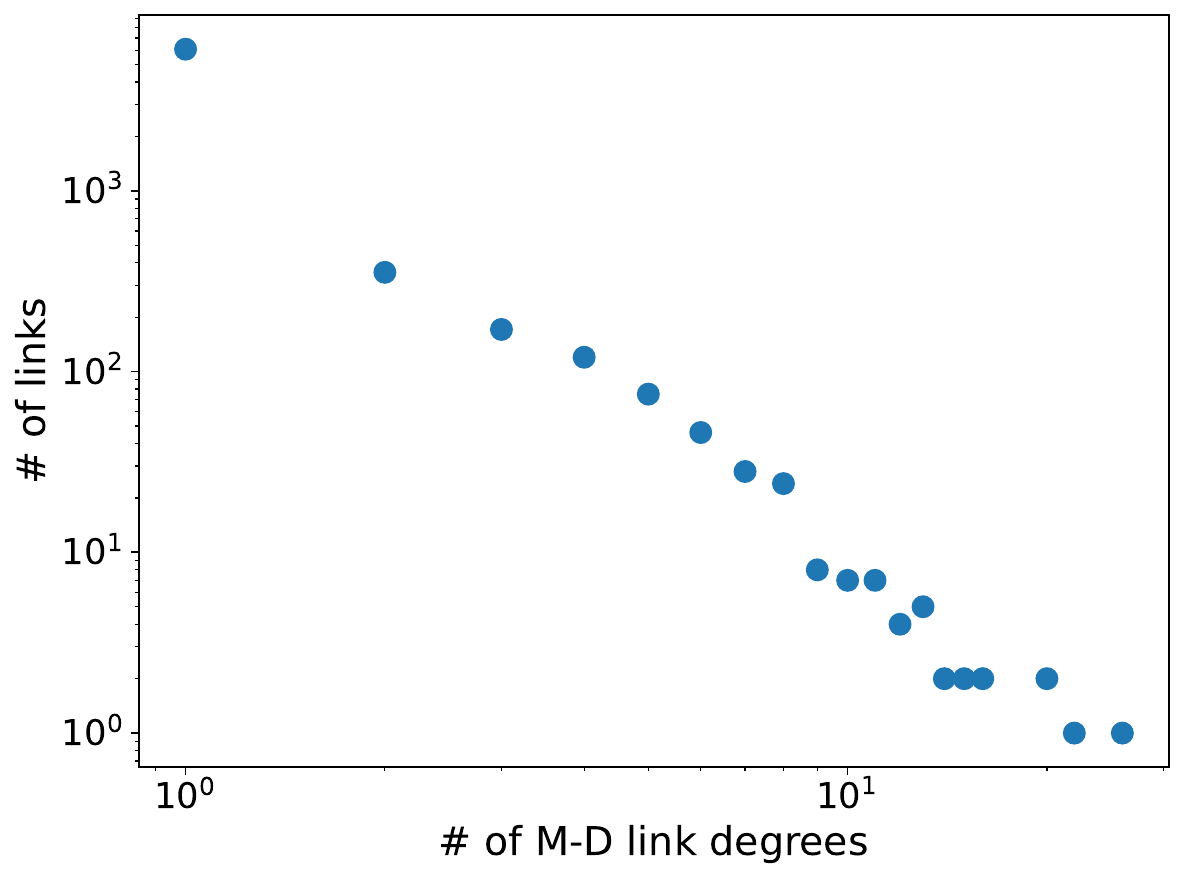}
        \subcaption{Movie-Director relation.}
        \vspace{0.25em}
    \end{minipage}\\
    \begin{minipage}[b]{0.33\linewidth}
        \centering
        \includegraphics[width=\linewidth]{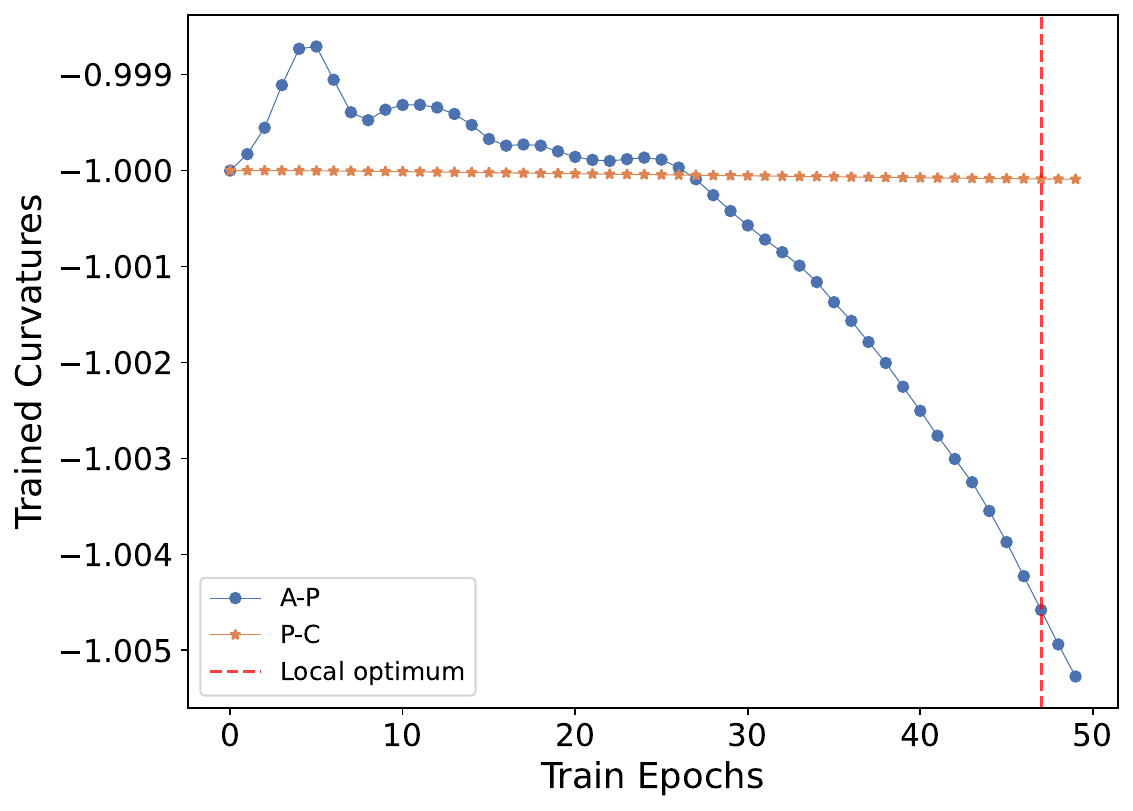}
        \subcaption{Results on DBLP dataset.}
    \end{minipage}
    \begin{minipage}[b]{0.32\linewidth}
        \centering
        \includegraphics[width=\linewidth]{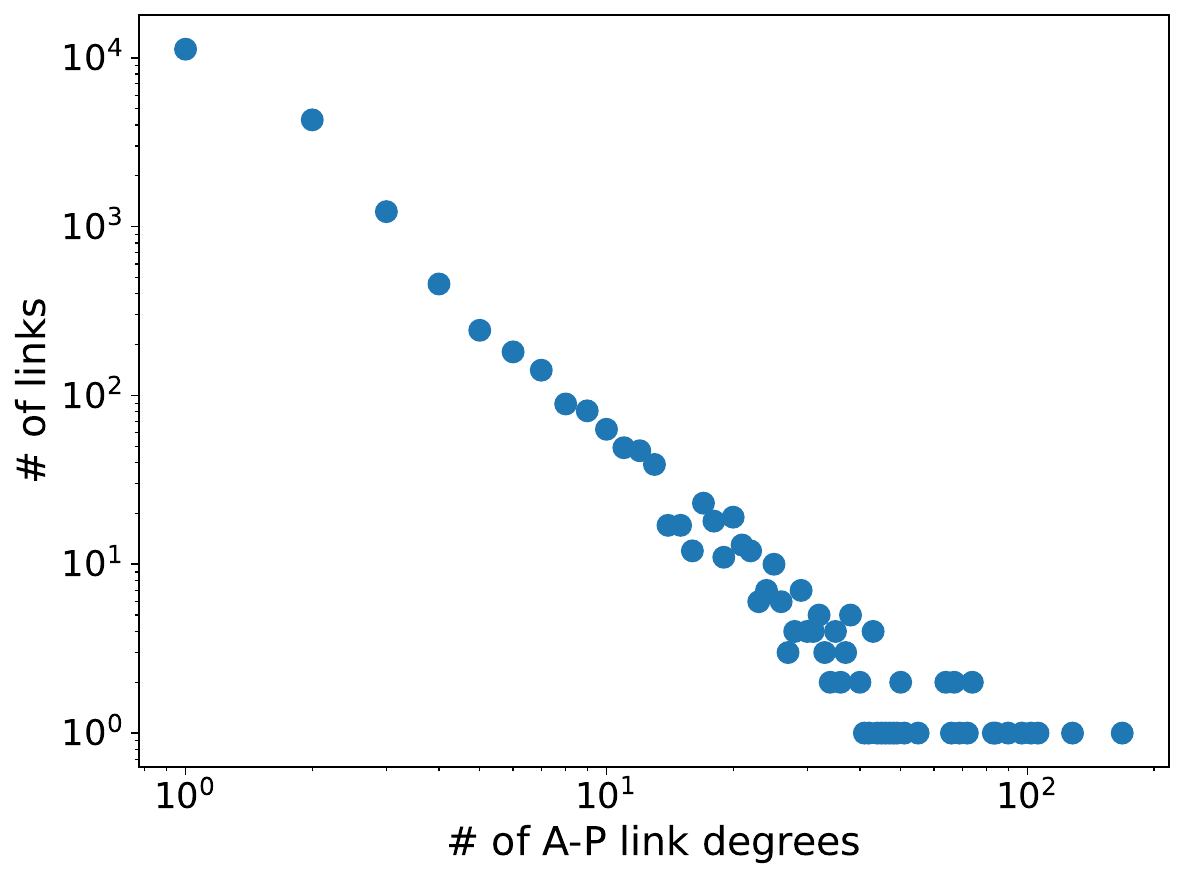}
        \subcaption{Author-Paper relation.}
    \end{minipage}
    \begin{minipage}[b]{0.32\linewidth}
        \centering
        \includegraphics[width=\linewidth]{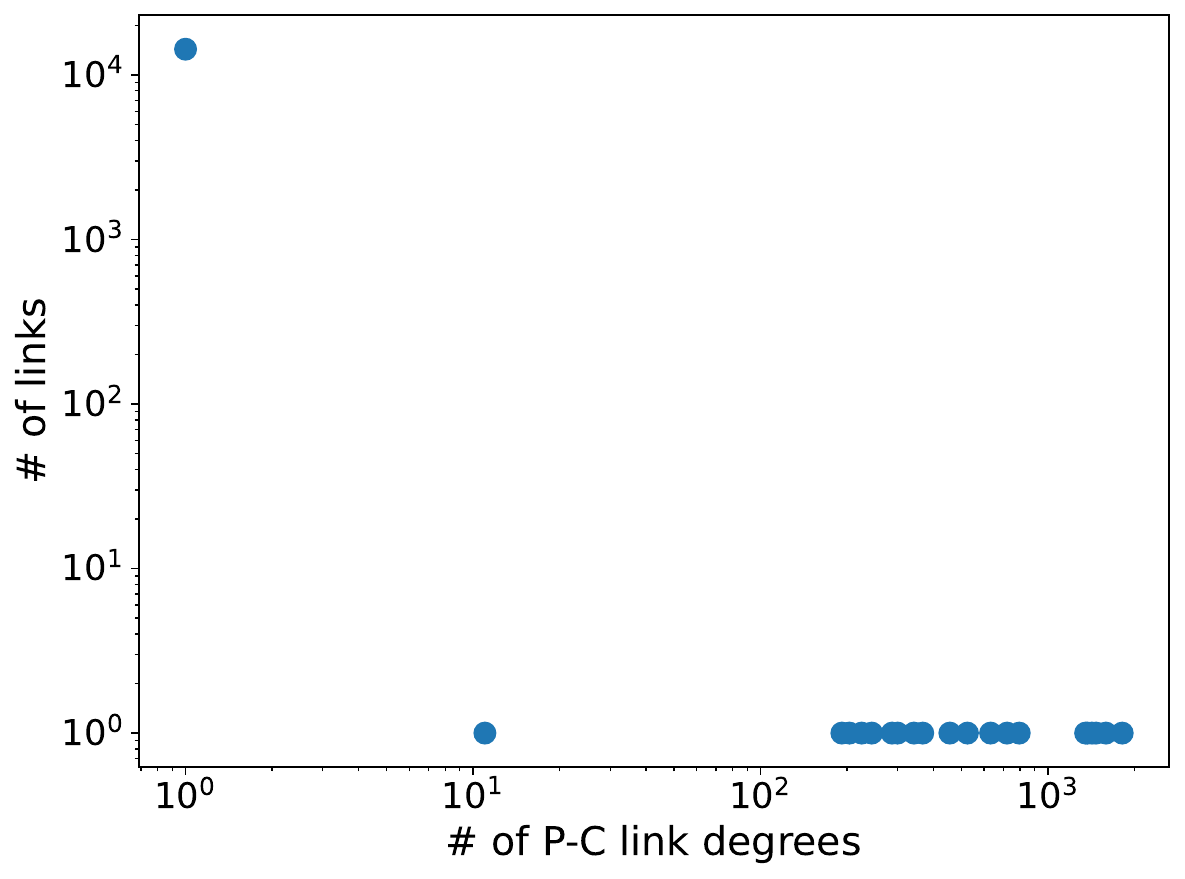}
        \subcaption{Paper-Conference relation.}
    \end{minipage}
    
\caption{Curvature variations of relation-specific hyperbolic spaces over train epochs on the IMDB and DBLP dataset.}
\label{figure:train_curvature}
\end{figure*}

\subsection{Hyperbolic Space Analysis (RQ3)}
First, to analyze the effectiveness of using multiple hyperbolic spaces for each relation type, we consider two variants. Figure~\ref{figure:space_analysis} represents the results of the hyperbolic space analysis and the details of each variant are as follows:
\begin{itemize}
    \item \textbf{Relation unawareness} is the variant that uses the input and output hyperbolic spaces as \algname{}. However, this model learns all relation-specific information within a single hyperbolic space rather than multiple relation-specific hyperbolic spaces.
    \item \textbf{Single hyperbolic space} is the variant that performs all operations for obtaining embeddings within a single hyperbolic space, without distinguishing between different relation-specific and input/output hyperbolic spaces.
\end{itemize}
Note that unlike the above variants, \algname{} uses both input/output hyperbolic spaces and relation-specific hyperbolic spaces for each relation type.
As shown in Figure~\ref{figure:space_analysis}, the results demonstrate the effectiveness of relation-specific hyperbolic spaces of \algname{} across all datasets.
Both in Micro-F1 and Macro-F1 variants show significant performance decrease. This results demonstrate that learning multiple relational distributions into a single hyperbolic space limits the ability to represent hierarchical structures corresponding to specific relation type effectively. In contrast, HypHGT, which learns relation-aware representations through multiple relation-specific hyperbolic spaces, effectively preserves the diversity of hierarchical structures corresponding to each relation type.

Second, Figure~\ref{figure:train_curvature} illustrates how the learnable curvatures of relation-specific hyperbolic spaces converge to their optimal values on the IMDB and DBLP datasets during \algname{} training. The number of training epochs is set to 50, and the vertical red dashed lines indicate the local optimal values achieved within 50 epochs. Figure~\ref{figure:train_curvature}(b), (c) and Figure~\ref{figure:train_curvature}(e), (f) show the link-degree distributions of each relation corresponding their underlying structures in IMDB and DBLP datasets, respectively. In the IMDB dataset, as shown in Figure~\ref{figure:train_curvature}(b) and (c), both Movie-Actor (M-A) and Movie-Director (M-D) relations shows similar power-law degree distributions, indicating comparable hierarchical structures. Accordingly, as shown in Figure~\ref{figure:train_curvature}(a), the curvatures of M-A and M-D relations converge to nearly similar values, demonstrating that \algname{} learns the curvature of each relation-specific hyperbolic space in accordance with its hierarchical structure.

In contrast, in the DBLP dataset, the Author-Paper (A-P) relation (Figure~\ref{figure:train_curvature}(e)) shows a power-law degree distribution, where as the Paper-Conference relation (Figure~\ref{figure:train_curvature}(f)) shows a flatter distribution with weak hierarchical properties. Consistent with this observation, as shown in Figure\ref{figure:train_curvature}(d), the curvature of the A-P relation decreases substantially to capture its hierarchical structures, while the curvature of P-C relation almost unchanged from its initial curvature. This result demonstrate that \algname{} adaptively adjusts the curvature of relation-specific hyperbolic space.
%to capture structural properties corresponding to specific relation. 
These results highlight that the learnable curvature in \algname{} is closely aligned with the structural properties of each relation rather than being uniform across all relations, and demonstrate that, through the proposed relation-specific hyperbolic spaces, \algname{} can effectively learn relation-specific representations that capture the distinct structural properties of each relation type.

\subsection{Scalability and Complexity Analysis (RQ4)}
To analyze the scalability of \algname{}, we generate a synthetic heterogeneous graph using the Barabasi-Albert model~\cite{barabasi1999emergence}, which captures the scale-free property commonly observed in real-world networks. First, we generate a power-law distributed homogeneous graph with the Brabasi-Albert model. Then, to give heterogeneity, we randomly assign node types to all nodes within this homogeneous graph, resulting in a heterogeneous graph. Finally, the generated synthetic heterogeneous graph consists of three node types \{A,B,C\} in a ratio of 6:3:1 and two link types \{AB, AC\}. The number of nodes is varied from 10K to 5M, allowing us to construct heterogeneous graphs of different scales and evaluate the time and memory consumption of the model under increasing graph scales. We set the batch size to 512 for training on large-scale synthetic heterogeneous graphs.

\begin{figure*}[t!]
\captionsetup[subfigure]{justification=centering}
\centering
    \begin{minipage}[b]{0.40\textwidth}
        \centering
        \includegraphics[width=\linewidth]{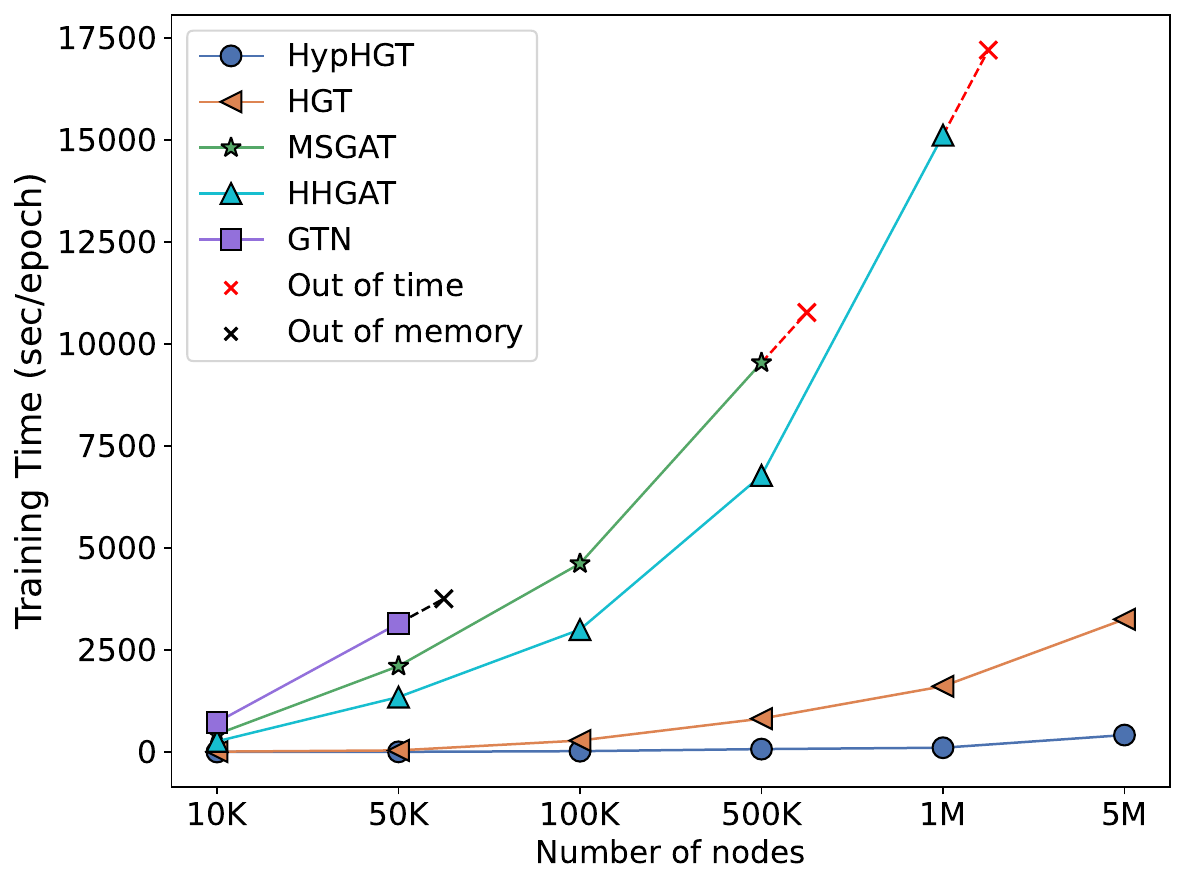}
        \subcaption{Time consumption for varying number of nodes.}
        \label{figure:scalability_time}
    \end{minipage}
    \begin{minipage}[b]{0.42\textwidth}
        \centering
        \includegraphics[width=\linewidth]{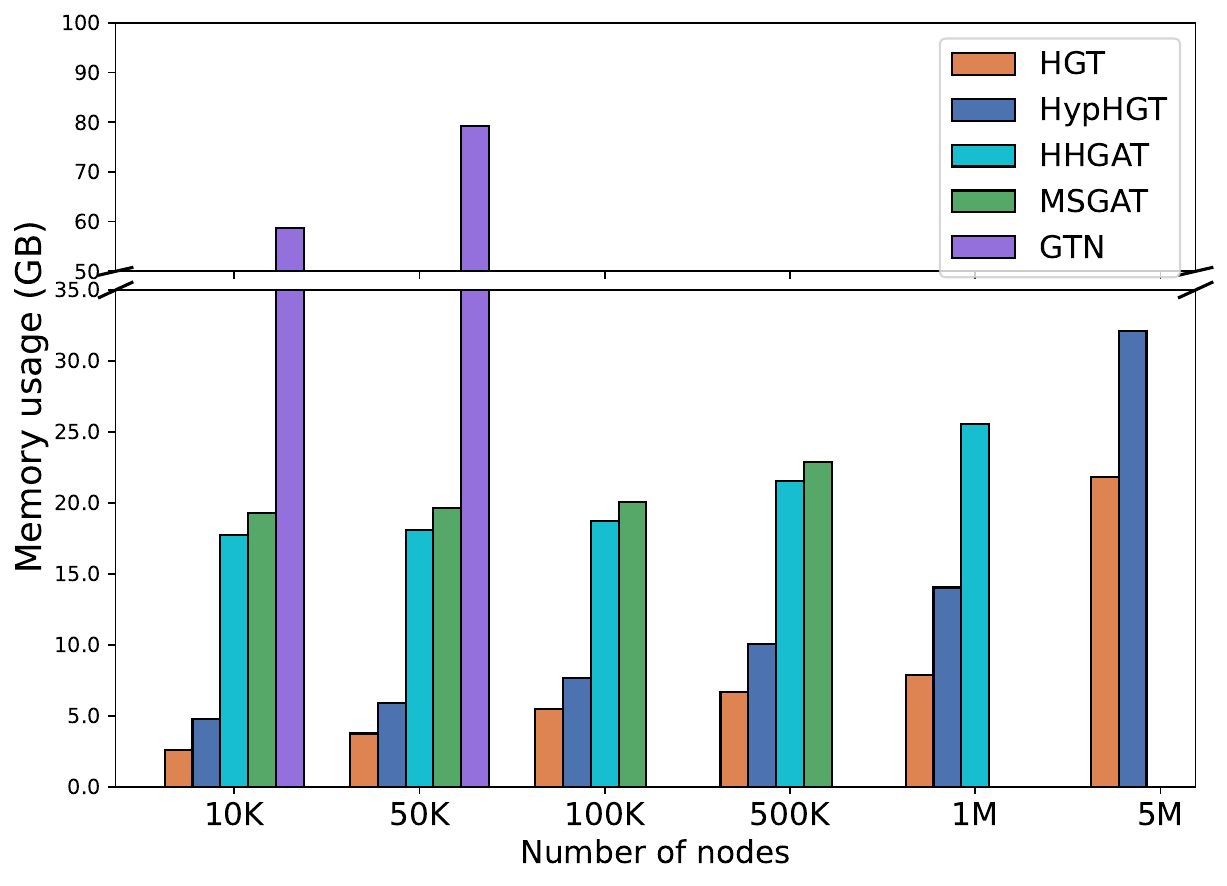}
        \subcaption{Memory consumption for varying number of nodes.}
        \label{figure:scalability_memory}
    \end{minipage}
    \iffalse
      \begin{minipage}[b]{0.315\linewidth}
        \centering
        \includegraphics[width=\linewidth]{Figures/sythetic_num_links.pdf}
        \subcaption{Number of links for varying number of nodes.}
        \label{figure:scalability_numlinks}
    \end{minipage}
    \fi
\caption{Scalability analysis on synthetic heterogeneous graphs.}
\label{figure:scalability}
\end{figure*}

\begin{figure}[t!]
\captionsetup[subfigure]{justification=centering}
\centering
\includegraphics[width=\columnwidth]{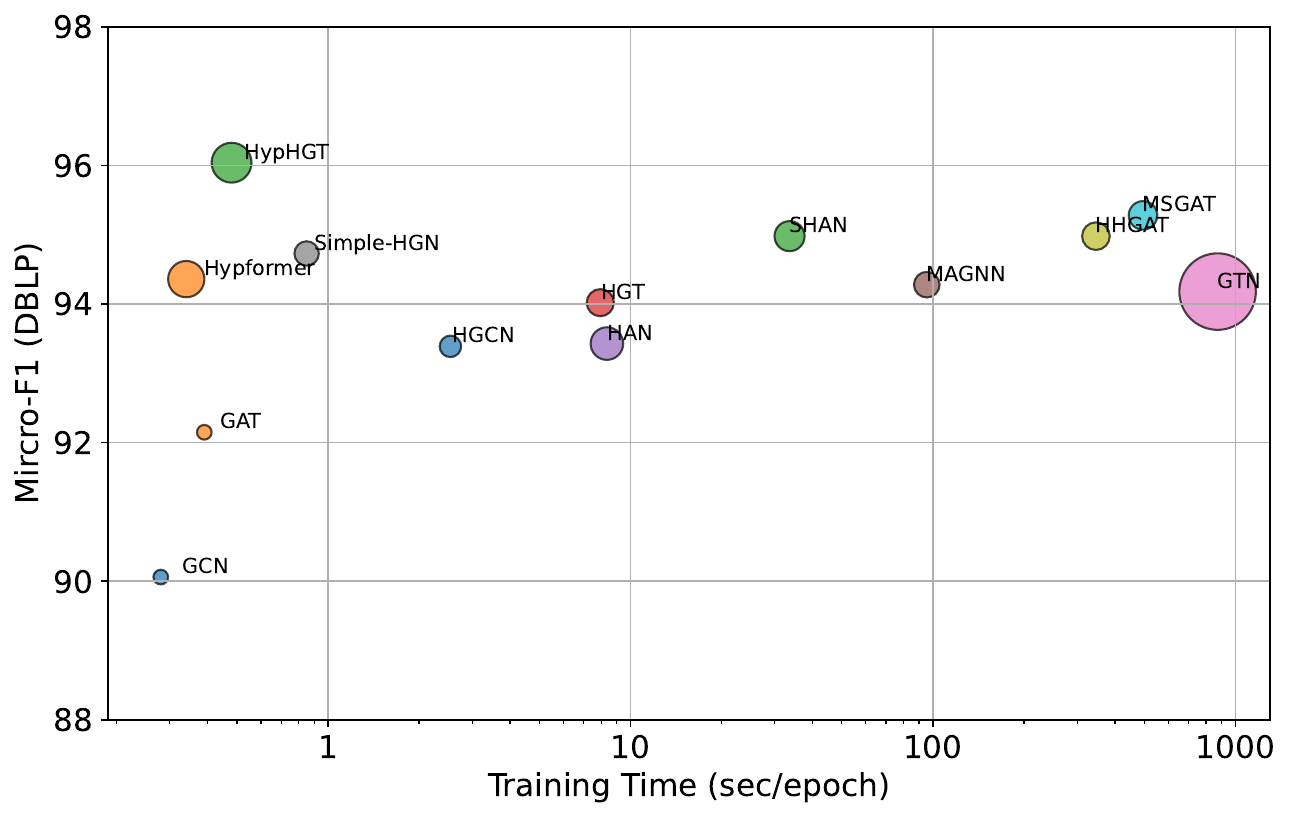}
\caption{Time and memory comparison for baselines on DBLP dataset. The area of the circles represents the relative memory consumption.}
\label{figure:benchmark}
\end{figure}

Figures~\ref{figure:scalability}(a) and \ref{figure:scalability}(b) illustrate the scalability analysis of \algname{} in terms of time and memory consumption, respectively, with respect to number of nodes in the synthetic heterogeneous graphs. In Figure~\ref{figure:scalability}(a), \algname{} shows near-linear growth in training time as the graph size increases from 10K to 5M nodes, while other baselines such as GTN, HHGAT, and MSGAT suffer from exponential increases in computation time, with several failing to complete training due to out-of-time or out-of-memory issues. In contrast, HGT, which was proposed to learn web-scale heterogeneous graphs, is able to scale effectively even as the graph size increases. Figure~\ref{figure:scalability}(b) further confirms that \algname{} maintains consistently low memory usage across all scales, while MSGAT and GTN consume a large amount of memory with increasing graph size. These empirical results are well aligned with the theoretical analysis in Section~\ref{sec:time_complex}: the near-linear training time validates the overall $O(N)$ complexity of \algname{} induced by the linear hyperbolic attention mechanism. Moreover, as shown in Figure~\ref{figure:benchmark}, \algname{} achieves high accuracy with significantly lower training time and memory consumption compared to other baselines, demonstrating its efficiency in learning real-world heterogeneous graphs.

\subsection{Hyperparameter Sensitivity Analysis}
\begin{figure*}[h!]
\captionsetup[subfigure]{justification=centering}
\centering
    \begin{minipage}[b]{0.3\linewidth}
        \centering
        \includegraphics[width=\linewidth]{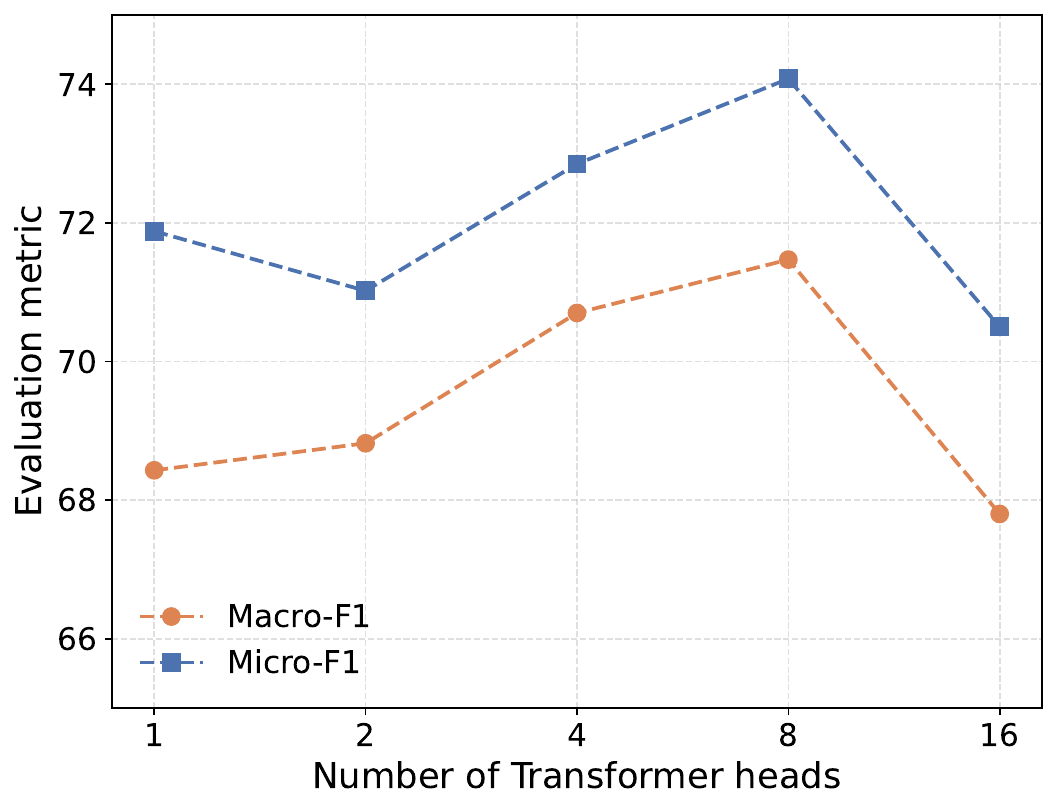}
        \subcaption{Number of transformer heads}
    \end{minipage}
    \begin{minipage}[b]{0.3\linewidth}
        \centering
        \includegraphics[width=\linewidth]{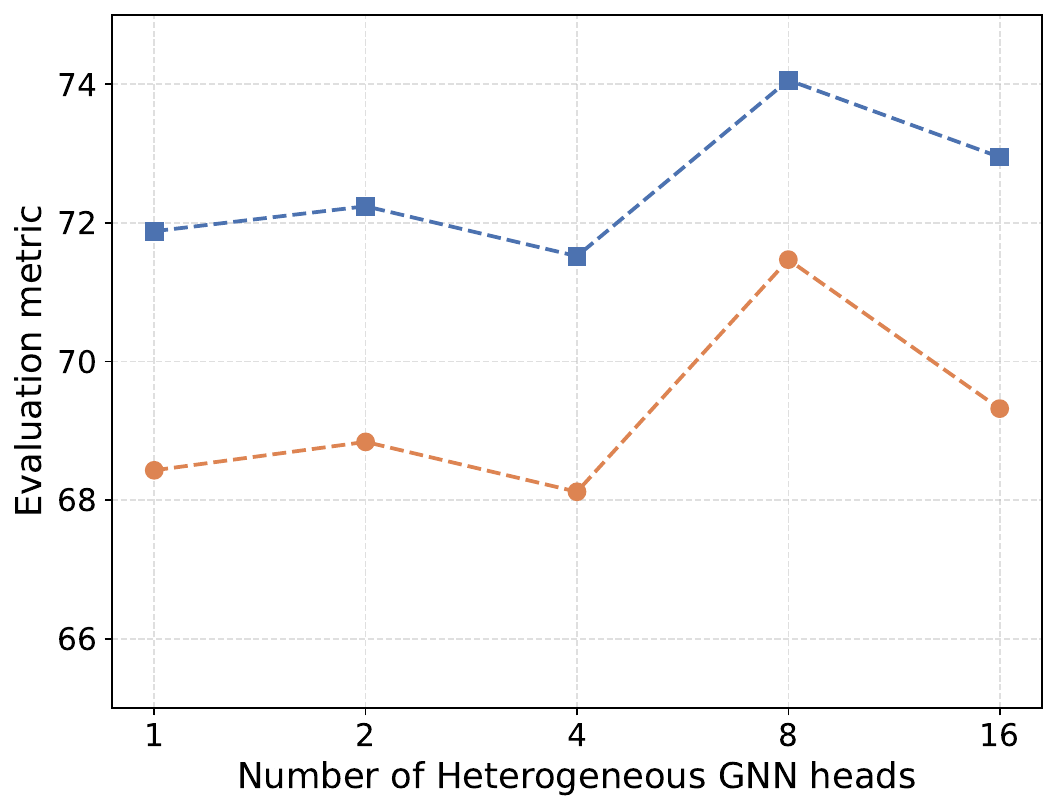}
        \subcaption{Number of heterogeneous GNN heads}
    \end{minipage}
    \begin{minipage}[b]{0.3\linewidth}
        \centering
        \includegraphics[width=\linewidth]{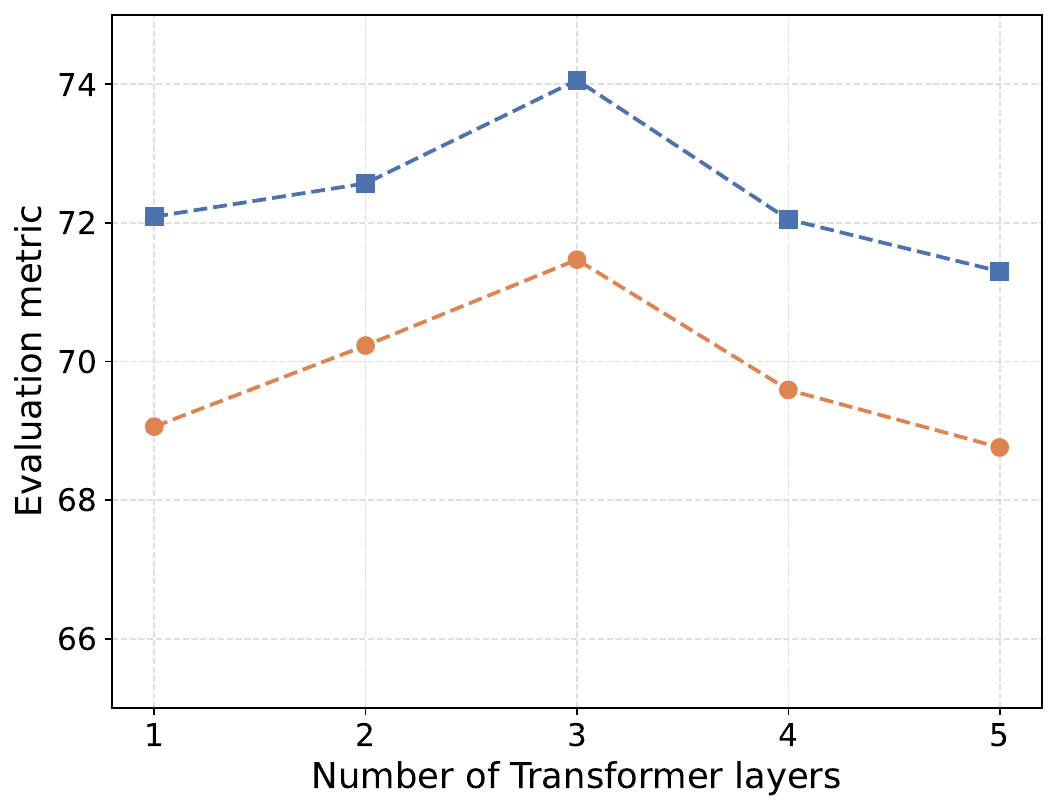}
        \subcaption{Number of transformer layers}
    \end{minipage}\\
    \begin{minipage}[b]{0.3\linewidth}
        \centering
        \includegraphics[width=\linewidth]{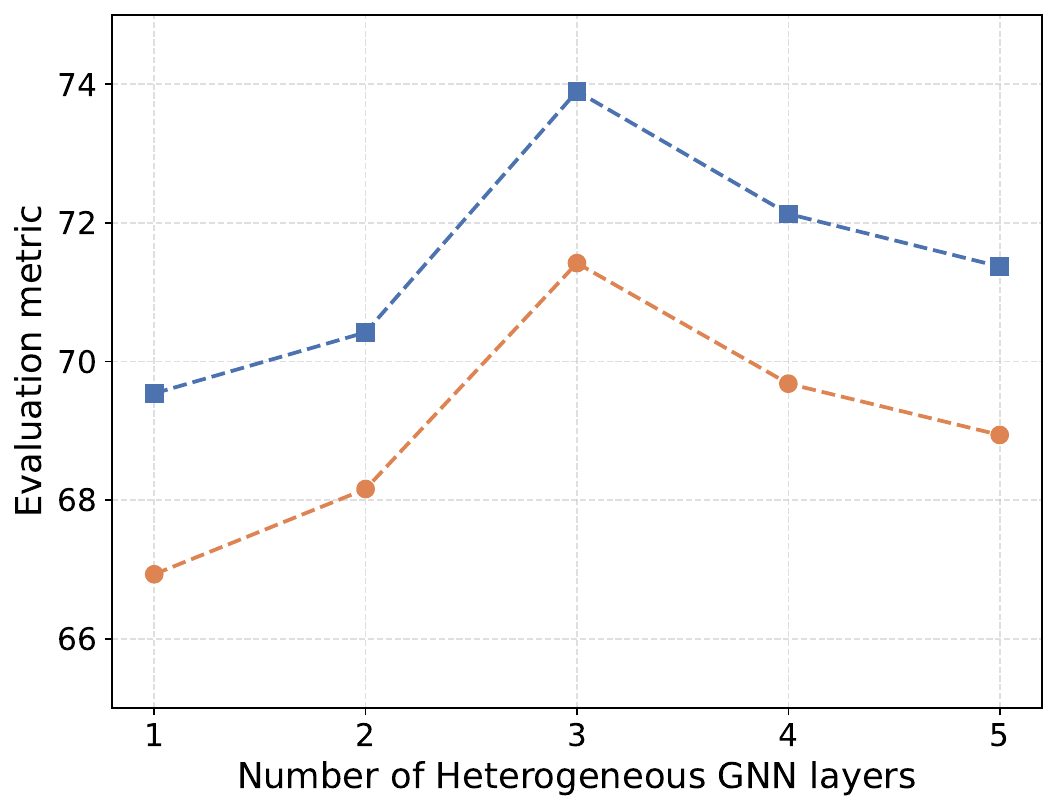}
        \subcaption{Number of heterogeneous GNN layers}
    \end{minipage}
    \begin{minipage}[b]{0.3\linewidth}
        \centering
        \includegraphics[width=\linewidth]{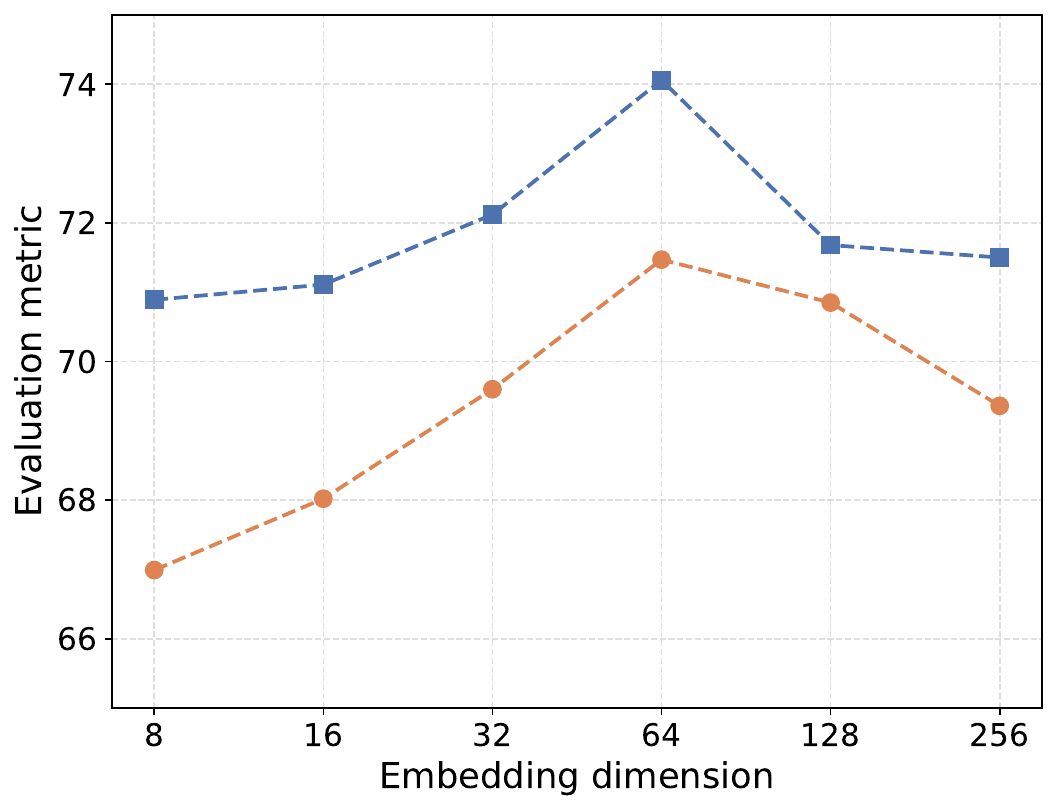}
        \subcaption{Dimension of node embedding}
    \end{minipage}
    \begin{minipage}[b]{0.3\linewidth}
        \centering
        \includegraphics[width=\linewidth]{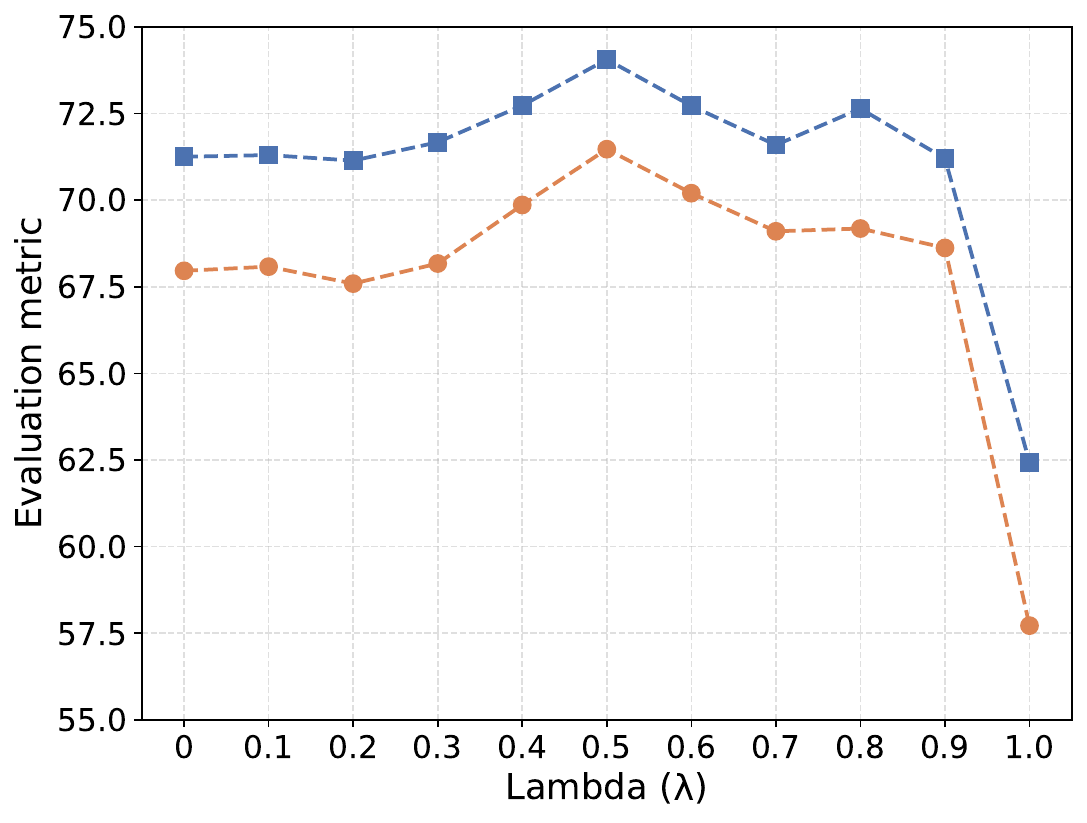}
        \subcaption{lambda ($\lambda$)}
    \end{minipage}
\caption{Hyperparameter sensitivity of \algname{}.}
\label{figure:hyperparams}
\end{figure*}

We investigate the impact of hyperparameters used in \algname{} and report the performance on node classification using the IMDB dataset. As shown in As shown in Figure~\ref{figure:hyperparams}(a), the performance increases as the number of hyperbolic heterogeneous graph transformer heads increases, reaching its peak at eight heads. However, when the number exceeds eight, the performance decreases. Similarly, in Figure~\ref{figure:hyperparams}(b), the performance improves as the number of heterogeneous GNN heads increases, peaking at eight heads. Beyond eight, the performance decreases. Next, as shown in Figure~\ref{figure:hyperparams}(c) and (d), increasing the number of hyperbolic heterogeneous graph transformer layers and heterogeneous GNNs layers enhances performance up to three. However, when the number of each layer increases beyond three, the performance starts to decrease. Figure~\ref{figure:hyperparams}(e) shows that, as the dimension of node embedding increases, the performance of \algname{} improves, reaching its best performance when the dimension is 64. However, beyond this peak, the performance decreases steadily. Figure~\ref{figure:hyperparams}(f) shows that a $\lambda$ value of 0.5 results in the optimal performance on the IMDB dataset. $\lambda$ controls the balance between the representations learned by the hyperbolic heterogeneous graph transformer and the heterogeneous GNNs. When $\lambda$ decreases, \algname{} relies more on the representations from the heterogeneous GNNs to obtain the final node embeddings. Conversely, as $\lambda$ increases, it relies more on the representations from the hyperbolic heterogeneous graph transformer. Notably, when $\lambda$ is set to 1, \algname{} depends solely on the hyperbolic heterogeneous graph transformer to obtain the node embeddings. Similar to the observation in the ablation study, this setting (i.e., $\lambda=1$) leads to a significant performance decrease, as it fails to capture the local neighbor information.

\section{Conclusion}
\label{sec:Conclusion}
In this paper, we proposed Hyperbolic Heterogeneous Graph Transformer (HypHGT), a novel model for efficient and effective heterogeneous graph representation learning in the hyperbolic space. Unlike previous hyperbolic heterogeneous embedding models that rely on tangent space and predefined metapath structures, \algname{} performs hyperbolic attention mechanism entirely within relation-specific hyperbolic spaces and avoids the requirement for dataset-dependent metapath definitions. Specifically, we proposes a linear hyperbolic heterogeneous attention mechanism that enables relation-aware representation learning in multiple hyperbolic spaces, significantly reducing time and memory consumption compared to previous hyperbolic heterogeneous graph embedding models. Furthermore, by integrating heterogeneous GNN layers with the hyperbolic transformer layers, \algname{} effectively captures both global hierarchical structures and neighbor information, leading to semantically rich node representations.

Comprehensive experiments on real-world and synthetic heterogeneous graphs demonstrated that \algname{} consistently outperforms state-of-the-art baselines in node classification tasks while maintaining superior efficiency. Ablation and hyperbolic space analysis confirmed that relation-specific hyperbolic spaces and joint learning of local and global information are key to achieving this improvement.

Potential avenues of future research include 1) extending \algname{} to dynamic heterogeneous graphs and its application to cross-domain scenarios where relational distributions change over time and 2) investigating curvature-adaptive training strategies for improving stability and interpretability of hyperbolic representations in heterogeneous graph learning.

\iffalse
\section*{Acknowledgment}
This work was supported by the National Research Foundation of Korea(NRF) grant funded by the Korea government(MSIT) (No. RS-2025-25435830).
\fi

\bibliographystyle{IEEEtran}
\bibliography{citations}

\end{document}